\documentclass{article}
\usepackage[T1]{fontenc}
\usepackage{loopreg_preprint,times}
\usepackage{amsmath,amssymb,amsthm,graphicx,booktabs,microtype,longtable}
\usepackage{colortbl}
\usepackage{hyperref,url}
\newcommand{\result}[1]{\ifcsname result@#1\endcsname\csname result@#1\endcsname\else\PackageError{LoopREG}{Missing result #1}{Rebuild the canonical data pipeline}\fi}

\newcommand{\ktr}{K_{\mathrm{train}}}

\newcommand{\Lrep}{\mathcal{L}_{\mathrm{rep}}}
\newtheorem{proposition}{Proposition}

\newcommand{\kinfer}{K_{\mathrm{infer}}}

\expandafter\def\csname result@SITbasefid\endcsname{35.25}
\expandafter\def\csname result@SITbasesfid\endcsname{6.65}
\expandafter\def\csname result@SITbaseis\endcsname{42.14}
\expandafter\def\csname result@SITbaseprecision\endcsname{0.530}
\expandafter\def\csname result@SITbaserecall\endcsname{0.650}
\expandafter\def\csname result@REGbasefid\endcsname{15.08}
\expandafter\def\csname result@REGbasesfid\endcsname{6.73}
\expandafter\def\csname result@REGbaseis\endcsname{95.53}
\expandafter\def\csname result@REGbaseprecision\endcsname{0.649}
\expandafter\def\csname result@REGbaserecall\endcsname{0.640}
\expandafter\def\csname result@SiTtwofid\endcsname{34.45}
\expandafter\def\csname result@SiTtwosfid\endcsname{6.43}
\expandafter\def\csname result@SiTtwois\endcsname{43.01}
\expandafter\def\csname result@SiTtwoprecision\endcsname{0.533}
\expandafter\def\csname result@SiTtworecall\endcsname{0.639}
\expandafter\def\csname result@SiTthreefid\endcsname{46.41}
\expandafter\def\csname result@SiTthreesfid\endcsname{10.01}
\expandafter\def\csname result@SiTthreeis\endcsname{34.28}
\expandafter\def\csname result@SiTthreeprecision\endcsname{0.464}
\expandafter\def\csname result@SiTthreerecall\endcsname{0.618}
\expandafter\def\csname result@REPAtwofid\endcsname{21.93}
\expandafter\def\csname result@REPAtwosfid\endcsname{6.37}
\expandafter\def\csname result@REPAtwois\endcsname{66.58}
\expandafter\def\csname result@REPAtwoprecision\endcsname{0.601}
\expandafter\def\csname result@REPAtworecall\endcsname{0.651}
\expandafter\def\csname result@REPAthreefid\endcsname{16.74}
\expandafter\def\csname result@REPAthreesfid\endcsname{6.68}
\expandafter\def\csname result@REPAthreeis\endcsname{84.67}
\expandafter\def\csname result@REPAthreeprecision\endcsname{0.629}
\expandafter\def\csname result@REPAthreerecall\endcsname{0.615}
\expandafter\def\csname result@REGtwofid\endcsname{14.71}
\expandafter\def\csname result@REGtwosfid\endcsname{6.76}
\expandafter\def\csname result@REGtwois\endcsname{97.13}
\expandafter\def\csname result@REGtwoprecision\endcsname{0.650}
\expandafter\def\csname result@REGtworecall\endcsname{0.631}
\expandafter\def\csname result@REGthreefid\endcsname{5.34}
\expandafter\def\csname result@REGthreesfid\endcsname{5.80}
\expandafter\def\csname result@REGthreeis\endcsname{185.98}
\expandafter\def\csname result@REGthreeprecision\endcsname{0.796}
\expandafter\def\csname result@REGthreerecall\endcsname{0.508}
\expandafter\def\csname result@XthreeMatchedfid\endcsname{13.75}
\expandafter\def\csname result@XthreeMatchedsfid\endcsname{6.69}
\expandafter\def\csname result@XthreeMatchedis\endcsname{100.82}
\expandafter\def\csname result@XthreeMatchedprecision\endcsname{0.658}
\expandafter\def\csname result@XthreeMatchedrecall\endcsname{0.634}
\expandafter\def\csname result@XthreeBestfid\endcsname{4.91}
\expandafter\def\csname result@XthreeBestsfid\endcsname{5.40}
\expandafter\def\csname result@XthreeBestis\endcsname{217.61}
\expandafter\def\csname result@XthreeBestprecision\endcsname{0.851}
\expandafter\def\csname result@XthreeBestrecall\endcsname{0.452}
\expandafter\def\csname result@Ltwofid\endcsname{4.23}
\expandafter\def\csname result@Ltwosfid\endcsname{5.09}
\expandafter\def\csname result@Ltwois\endcsname{170.36}
\expandafter\def\csname result@Ltwoprecision\endcsname{0.752}
\expandafter\def\csname result@Ltworecall\endcsname{0.642}
\expandafter\def\csname result@Lthreefid\endcsname{3.21}
\expandafter\def\csname result@Lthreesfid\endcsname{4.44}
\expandafter\def\csname result@Lthreeis\endcsname{262.66}
\expandafter\def\csname result@Lthreeprecision\endcsname{0.859}
\expandafter\def\csname result@Lthreerecall\endcsname{0.520}
\expandafter\def\csname result@MainXLtwofid\endcsname{3.38}
\expandafter\def\csname result@MainXLtwosfid\endcsname{4.85}
\expandafter\def\csname result@MainXLtwois\endcsname{184.71}
\expandafter\def\csname result@MainXLtwoprecision\endcsname{0.758}
\expandafter\def\csname result@MainXLtworecall\endcsname{0.635}
\expandafter\def\csname result@MainXLthreefid\endcsname{2.84}
\expandafter\def\csname result@MainXLthreesfid\endcsname{4.64}
\expandafter\def\csname result@MainXLthreeis\endcsname{254.70}
\expandafter\def\csname result@MainXLthreeprecision\endcsname{0.839}
\expandafter\def\csname result@MainXLthreerecall\endcsname{0.549}
\expandafter\def\csname result@PublishedREPAB2fid\endcsname{24.40}
\expandafter\def\csname result@PublishedREPAB2sfid\endcsname{6.40}
\expandafter\def\csname result@PublishedREPAB2is\endcsname{59.90}
\expandafter\def\csname result@PublishedREPAB2precision\endcsname{0.590}
\expandafter\def\csname result@PublishedREPAB2recall\endcsname{0.650}
\expandafter\def\csname result@PublishedSiTL2fid\endcsname{18.80}
\expandafter\def\csname result@PublishedSiTL2sfid\endcsname{5.29}
\expandafter\def\csname result@PublishedSiTL2is\endcsname{72.00}
\expandafter\def\csname result@PublishedSiTL2precision\endcsname{0.640}
\expandafter\def\csname result@PublishedSiTL2recall\endcsname{0.640}
\expandafter\def\csname result@PublishedREPAL2fid\endcsname{10.00}
\expandafter\def\csname result@PublishedREPAL2sfid\endcsname{5.20}
\expandafter\def\csname result@PublishedREPAL2is\endcsname{109.20}
\expandafter\def\csname result@PublishedREPAL2precision\endcsname{0.690}
\expandafter\def\csname result@PublishedREPAL2recall\endcsname{0.650}
\expandafter\def\csname result@PublishedREGL2fid\endcsname{4.60}
\expandafter\def\csname result@PublishedREGL2sfid\endcsname{5.21}
\expandafter\def\csname result@PublishedREGL2is\endcsname{167.60}
\expandafter\def\csname result@PublishedREGL2precision\endcsname{0.750}
\expandafter\def\csname result@PublishedREGL2recall\endcsname{0.630}
\expandafter\def\csname result@PublishedSiTXL2fid\endcsname{8.30}
\expandafter\def\csname result@PublishedSiTXL2sfid\endcsname{6.32}
\expandafter\def\csname result@PublishedSiTXL2is\endcsname{131.70}
\expandafter\def\csname result@PublishedSiTXL2precision\endcsname{0.680}
\expandafter\def\csname result@PublishedSiTXL2recall\endcsname{0.670}
\expandafter\def\csname result@PublishedREPAXL2fid\endcsname{5.90}
\expandafter\def\csname result@PublishedREPAXL2sfid\endcsname{5.73}
\expandafter\def\csname result@PublishedREPAXL2is\endcsname{157.80}
\expandafter\def\csname result@PublishedREPAXL2precision\endcsname{0.700}
\expandafter\def\csname result@PublishedREPAXL2recall\endcsname{0.690}
\expandafter\def\csname result@PublishedREGXL2fid\endcsname{3.40}
\expandafter\def\csname result@PublishedREGXL2sfid\endcsname{4.87}
\expandafter\def\csname result@PublishedREGXL2is\endcsname{184.10}
\expandafter\def\csname result@PublishedREGXL2precision\endcsname{0.760}
\expandafter\def\csname result@PublishedREGXL2recall\endcsname{0.640}
\expandafter\def\csname result@REPAGain\endcsname{23.64}
\expandafter\def\csname result@REGGain\endcsname{63.67}
\expandafter\def\csname result@XLGain\endcsname{16.10}
\expandafter\def\csname result@BaselineToX3Gain\endcsname{67.41}
\expandafter\def\csname result@Train4K6FID\endcsname{5.19}
\expandafter\def\csname result@HorizonSlope\endcsname{1.5517}
\expandafter\def\csname result@D4B2K2\endcsname{13.92}
\expandafter\def\csname result@D4B2K3\endcsname{32.56}
\expandafter\def\csname result@D4B2K4\endcsname{58.85}
\expandafter\def\csname result@D4B2Gain\endcsname{-133.93}
\expandafter\def\csname result@D4B2Recall\endcsname{0.676}
\expandafter\def\csname result@D4B4K2\endcsname{14.71}
\expandafter\def\csname result@D4B4K3\endcsname{5.34}
\expandafter\def\csname result@D4B4K4\endcsname{6.20}
\expandafter\def\csname result@D4B4Gain\endcsname{63.67}
\expandafter\def\csname result@D4B4Recall\endcsname{0.508}
\expandafter\def\csname result@D4B6K2\endcsname{14.53}
\expandafter\def\csname result@D4B6K3\endcsname{31.11}
\expandafter\def\csname result@D4B6K4\endcsname{69.58}
\expandafter\def\csname result@D4B6Gain\endcsname{-114.13}
\expandafter\def\csname result@D4B6Recall\endcsname{0.611}
\expandafter\def\csname result@D4B8K2\endcsname{14.18}
\expandafter\def\csname result@D4B8K3\endcsname{29.97}
\expandafter\def\csname result@D4B8K4\endcsname{59.55}
\expandafter\def\csname result@D4B8Gain\endcsname{-111.33}
\expandafter\def\csname result@D4B8Recall\endcsname{0.618}
\expandafter\def\csname result@D2B2K2\endcsname{15.61}
\expandafter\def\csname result@D2B2K3\endcsname{7.49}
\expandafter\def\csname result@D2B2K4\endcsname{10.25}
\expandafter\def\csname result@D2B2Gain\endcsname{51.99}
\expandafter\def\csname result@D2B2Recall\endcsname{0.544}
\expandafter\def\csname result@D6B6K2\endcsname{14.81}
\expandafter\def\csname result@D6B6K3\endcsname{5.98}
\expandafter\def\csname result@D6B6K4\endcsname{7.78}
\expandafter\def\csname result@D6B6Gain\endcsname{59.61}
\expandafter\def\csname result@D6B6Recall\endcsname{0.528}
\expandafter\def\csname result@D8B8K2\endcsname{16.54}
\expandafter\def\csname result@D8B8K3\endcsname{12.88}
\expandafter\def\csname result@D8B8K4\endcsname{28.32}
\expandafter\def\csname result@D8B8Gain\endcsname{22.16}
\expandafter\def\csname result@D8B8Recall\endcsname{0.541}
\expandafter\def\csname result@Adaptivefid\endcsname{7.25}
\expandafter\def\csname result@Adaptiverecall\endcsname{0.566}
\expandafter\def\csname result@AdaptiveMeanK\endcsname{2.416}
\expandafter\def\csname result@AdaptiveCoreSavings\endcsname{19.47}
\expandafter\def\csname result@AdaptiveBlockSavings\endcsname{4.17}
\expandafter\def\csname result@UnsharedK2\endcsname{14.15}
\expandafter\def\csname result@UnsharedK4\endcsname{12.27}
\expandafter\def\csname result@Random123fid\endcsname{11.68}
\expandafter\def\csname result@Random123recall\endcsname{0.640}
\expandafter\def\csname result@Random123infer_k\endcsname{5}
\expandafter\def\csname result@LoopSiTx2NMSE2\endcsname{0.46074}
\expandafter\def\csname result@LoopREGx2NMSE2\endcsname{0.45706}
\expandafter\def\csname result@LoopREGx2NMSE3\endcsname{0.45898}
\expandafter\def\csname result@LoopREGx3NMSE3\endcsname{0.45684}
\expandafter\def\csname result@LoopREGx3NMSE5\endcsname{0.46031}
\expandafter\def\csname result@UnsharedThreeK3fid\endcsname{13.32}
\expandafter\def\csname result@UnsharedThreeK6fid\endcsname{40.91}
\expandafter\def\csname result@XLtwofid\endcsname{3.3823}
\expandafter\def\csname result@XLtwosfid\endcsname{4.8470}
\expandafter\def\csname result@XLtworecall\endcsname{0.6347}
\expandafter\def\csname result@XLthreefid\endcsname{2.8376}
\expandafter\def\csname result@XLthreerecall\endcsname{0.5490}
\expandafter\def\csname result@XLearlyfid\endcsname{3.3675}
\expandafter\def\csname result@XLearlysfid\endcsname{4.5372}
\expandafter\def\csname result@XLearlyrecall\endcsname{0.5180}
\expandafter\def\csname result@XLLocalStepsSaved\endcsname{45}

\expandafter\def\csname result@CausalTrain2Seed2BDGain\endcsname{29.1}
\expandafter\def\csname result@CausalTrain2Seed3BDGain\endcsname{28.4}
\expandafter\def\csname result@CausalTrain3Seed2BDGain\endcsname{29.6}
\expandafter\def\csname result@CausalTrain3Seed3BDGain\endcsname{29.0}

\expandafter\def\csname result@R4BelowFloor\endcsname{16}
\expandafter\def\csname result@R5MeanRegret\endcsname{8.744}

\expandafter\def\csname result@Refine2PatchCKABestK\endcsname{3}
\expandafter\def\csname result@Refine3PatchCKABestK\endcsname{5}

\title{LoopREG: Representation Supervision Enables\\Test-Time Recurrent Depth in Diffusion Transformers}
\author{\makebox[\textwidth][c]{\begin{minipage}{1.08\textwidth}\centering
\setlength{\tabcolsep}{0pt}
\begin{tabular}{@{}cccc@{}}
\begin{minipage}[t]{.25\textwidth}\centering
{\fontsize{9}{10.5}\selectfont \href{https://orcid.org/0009-0006-9720-7156}{Zipeng Wang}\textsuperscript{1,2}\endgraf}
\vspace{1pt}{\fontsize{7.8}{9}\selectfont
\href{mailto:d250232003@stu.cqupt.edu.cn}{d250232003@stu.cqupt.edu.cn}\endgraf
}\end{minipage} & \begin{minipage}[t]{.25\textwidth}\centering
{\fontsize{9}{10.5}\selectfont Anping Xiong\textsuperscript{1}\endgraf}
\vspace{1pt}{\fontsize{7.8}{9}\selectfont
\href{mailto:xiongap@cqupt.edu.cn}{xiongap@cqupt.edu.cn}\endgraf
}\end{minipage} & \begin{minipage}[t]{.25\textwidth}\centering
{\fontsize{9}{10.5}\selectfont Jing Zhang\textsuperscript{4}\endgraf}
\vspace{1pt}{\fontsize{7.8}{9}\selectfont
\href{mailto:jingzh007@alu.uestc.edu.cn}{jingzh007@alu.uestc.edu.cn}\endgraf
}\end{minipage} & \begin{minipage}[t]{.25\textwidth}\centering
{\fontsize{9}{10.5}\selectfont Xiaoya Zhu\textsuperscript{2}\endgraf}
\vspace{1pt}{\fontsize{7.8}{9}\selectfont
\href{mailto:zhuxy172@chinaunicom.cn}{zhuxy172@chinaunicom.cn}\endgraf
}\end{minipage}\\
\noalign{\vskip 3pt}
\begin{minipage}[t]{.25\textwidth}\centering
{\fontsize{9}{10.5}\selectfont Yi Jiang\textsuperscript{1,*}\endgraf}
\vspace{1pt}{\fontsize{7.8}{9}\selectfont
\href{mailto:jiangyi@cqupt.edu.cn}{jiangyi@cqupt.edu.cn}\endgraf
}\end{minipage} & \begin{minipage}[t]{.25\textwidth}\centering
{\fontsize{9}{10.5}\selectfont Yanbing Xiao\textsuperscript{1}\endgraf}
\vspace{1pt}{\fontsize{7.8}{9}\selectfont
\href{mailto:d240201045@stu.cqupt.edu.cn}{d240201045@stu.cqupt.edu.cn}\endgraf
}\end{minipage} & \begin{minipage}[t]{.25\textwidth}\centering
{\fontsize{9}{10.5}\selectfont Zige Zeng\textsuperscript{1}\endgraf}
\vspace{1pt}{\fontsize{7.8}{9}\selectfont
\href{mailto:d250101001@stu.cqupt.edu.cn}{d250101001@stu.cqupt.edu.cn}\endgraf
}\end{minipage} & \begin{minipage}[t]{.25\textwidth}\centering
{\fontsize{9}{10.5}\selectfont \href{https://orcid.org/0009-0000-6796-6043}{Zezhou Chen}\textsuperscript{2}\endgraf}
\vspace{1pt}{\fontsize{7.8}{9}\selectfont
\href{mailto:chenzz51@chinaunicom.cn}{chenzz51@chinaunicom.cn}\endgraf
}\end{minipage}\\
\noalign{\vskip 3pt}
\begin{minipage}[t]{.25\textwidth}\centering
{\fontsize{9}{10.5}\selectfont \href{https://orcid.org/0009-0003-2492-403X}{Xiang Liu}\textsuperscript{2}\endgraf}
\vspace{1pt}{\fontsize{7.8}{9}\selectfont
\href{mailto:liux750@chinaunicom.cn}{liux750@chinaunicom.cn}\endgraf
}\end{minipage} & \begin{minipage}[t]{.25\textwidth}\centering
{\fontsize{9}{10.5}\selectfont \href{https://orcid.org/0009-0009-6697-3220}{Huan Hu}\textsuperscript{2}\endgraf}
\vspace{1pt}{\fontsize{7.8}{9}\selectfont
\href{mailto:huh30@chinaunicom.cn}{huh30@chinaunicom.cn}\endgraf
}\end{minipage} & \begin{minipage}[t]{.25\textwidth}\centering
{\fontsize{9}{10.5}\selectfont \href{https://orcid.org/0009-0003-2430-6944}{Ping Chen}\textsuperscript{2}\endgraf}
\vspace{1pt}{\fontsize{7.8}{9}\selectfont
\href{mailto:chenp181@chinaunicom.cn}{chenp181@chinaunicom.cn}\endgraf
\href{mailto:redcping@gmail.com}{redcping@gmail.com}\endgraf
}\end{minipage} & \begin{minipage}[t]{.25\textwidth}\centering
{\fontsize{9}{10.5}\selectfont Kai Wang\textsuperscript{2}\endgraf}
\vspace{1pt}{\fontsize{7.8}{9}\selectfont
\href{mailto:wangk115@chinaunicom.cn}{wangk115@chinaunicom.cn}\endgraf
}\end{minipage}\\
\noalign{\vskip 3pt}
\begin{minipage}[t]{.25\textwidth}\centering
{\fontsize{9}{10.5}\selectfont Zhaoxiang Liu\textsuperscript{2}\endgraf}
\vspace{1pt}{\fontsize{7.8}{9}\selectfont
\href{mailto:liuzx178@chinaunicom.cn}{liuzx178@chinaunicom.cn}\endgraf
}\end{minipage} & \begin{minipage}[t]{.25\textwidth}\centering
{\fontsize{9}{10.5}\selectfont Shiguo Lian\textsuperscript{2,*}\endgraf}
\vspace{1pt}{\fontsize{7.8}{9}\selectfont
\href{mailto:liansg@chinaunicom.cn}{liansg@chinaunicom.cn}\endgraf
}\end{minipage} & \begin{minipage}[t]{.25\textwidth}\centering
{\fontsize{9}{10.5}\selectfont Ye LV\textsuperscript{3,*}\endgraf}
\vspace{1pt}{\fontsize{7.8}{9}\selectfont
\href{mailto:lvy30@chinaunicom.cn}{lvy30@chinaunicom.cn}\endgraf
}\end{minipage} & \begin{minipage}[t]{.25\textwidth}\centering
{\fontsize{9}{10.5}\selectfont Fang Zhao\textsuperscript{2}\endgraf}
\vspace{1pt}{\fontsize{7.8}{9}\selectfont
\href{mailto:zhaof50@chinaunicom.cn}{zhaof50@chinaunicom.cn}\endgraf
}\end{minipage}\\
\noalign{\vskip 3pt}
\begin{minipage}[t]{.25\textwidth}\centering
{\fontsize{9}{10.5}\selectfont Huanlin Gao\textsuperscript{2}\endgraf}
\vspace{1pt}{\fontsize{7.8}{9}\selectfont
\href{mailto:ghl@nwafu.edu.cn}{ghl@nwafu.edu.cn}\endgraf
}\end{minipage} & \begin{minipage}[t]{.25\textwidth}\centering
{\fontsize{9}{10.5}\selectfont Zhengrui Huang\textsuperscript{1}\endgraf}
\vspace{1pt}{\fontsize{7.8}{9}\selectfont
\href{mailto:d250201016@stu.cqupt.edu.cn}{d250201016@stu.cqupt.edu.cn}\endgraf
}\end{minipage} & \begin{minipage}[t]{.25\textwidth}\centering
{\fontsize{9}{10.5}\selectfont Bo Shen\textsuperscript{2}\endgraf}
\vspace{1pt}{\fontsize{7.8}{9}\selectfont
\href{mailto:shenb15@chinaunicom.cn}{shenb15@chinaunicom.cn}\endgraf
}\end{minipage} & \begin{minipage}[t]{.25\textwidth}\centering
{\fontsize{9}{10.5}\selectfont Yuren You\textsuperscript{2}\endgraf}
\vspace{1pt}{\fontsize{7.8}{9}\selectfont
\href{mailto:youyr@chinaunicom.cn}{youyr@chinaunicom.cn}\endgraf
}\end{minipage}\\
\end{tabular}
\endgraf\vspace{2pt}
{\normalsize
\begin{tabular}{@{}cc@{}}
\begin{minipage}[t]{.5\textwidth}\centering\fontsize{8}{9.5}\selectfont\raisebox{.5ex}{\fontsize{6}{6}\selectfont 1}Chongqing University of Posts and Telecommunications\end{minipage} & \begin{minipage}[t]{.5\textwidth}\centering\fontsize{8}{9.5}\selectfont\raisebox{.5ex}{\fontsize{6}{6}\selectfont 2}China United Network Communications Group Co., Ltd.\end{minipage}\\[2pt]
\begin{minipage}[t]{.5\textwidth}\centering\fontsize{8}{9.5}\selectfont\raisebox{.5ex}{\fontsize{6}{6}\selectfont 3}Dongbei University of Finance and Economics\end{minipage} & \begin{minipage}[t]{.5\textwidth}\centering\fontsize{8}{9.5}\selectfont\raisebox{.5ex}{\fontsize{6}{6}\selectfont 4}University of Electronic Science and Technology of China\end{minipage}\\[2pt]
\end{tabular}\endgraf\vspace{3pt}
\fontsize{8}{9.5}\selectfont * Corresponding authors (in order): Yi Jiang, Ye LV, Shiguo Lian.
}\end{minipage}}}

\hypersetup{pdftitle={LoopREG: Representation Supervision Enables Test-Time Recurrent Depth in Diffusion Transformers},pdfauthor={Zipeng Wang, Anping Xiong, Jing Zhang, Xiaoya Zhu, Yi Jiang, Yanbing Xiao, Zige Zeng, Zezhou Chen, Xiang Liu, Huan Hu, Ping Chen, Kai Wang, Zhaoxiang Liu, Shiguo Lian, Ye LV, Fang Zhao, Huanlin Gao, Zhengrui Huang, Bo Shen, Yuren You},pdfkeywords={diffusion transformers, representation supervision, recurrent depth, image generation},hidelinks}
\begin{document}
\raggedbottom
\maketitle
\begin{abstract}
Reusing Transformer blocks offers a way to increase diffusion-model computation at test time, but naive recurrence can degrade generation beyond its training depth. We introduce LoopREG, a representation-supervised recurrent diffusion Transformer that repeatedly applies a shared block at the state supervised by a pretrained visual encoder. Alignment is applied after the final recurrent visit during training; inference extends the same learned operator to additional visits. On ImageNet $256\times256$, LoopREG reduces SiT-B/2 FID from the local REG baseline's \result{REGbasefid} to \result{XthreeBestfid} at 400K training steps, a \result{BaselineToX3Gain}\% reduction. Within a checkpoint trained with two visits, one extra inference visit improves FID from \result{REGtwofid} to \result{REGthreefid} with unchanged weights. The benefit persists on SiT-L/2 and SiT-XL/2. Controlled interventions show that useful recurrence follows the location of representation supervision, while REPA alignment produces a smaller extrapolation benefit. These results make representation-supervised recurrence a practical route to improving diffusion generation through additional inference depth.
\end{abstract}
\begin{figure}[!ht]
\centering\includegraphics[width=.96\linewidth]{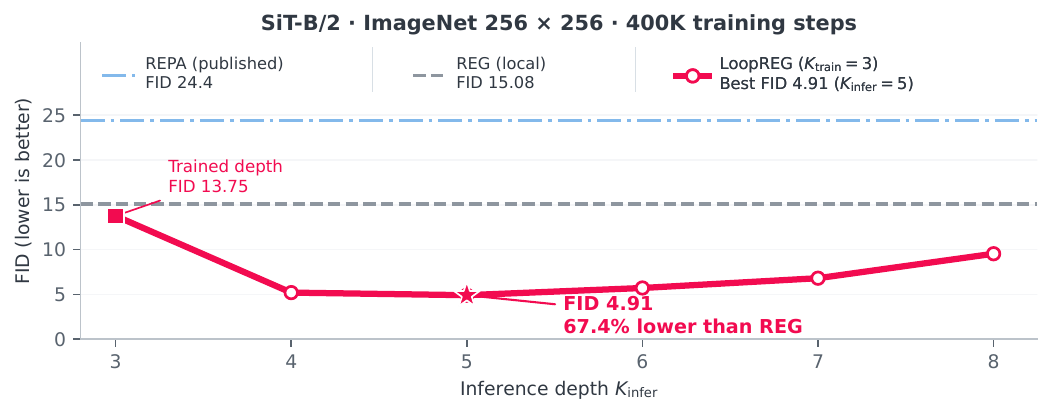}
\caption{\textbf{LoopREG reduces REG's FID by \result{BaselineToX3Gain}\% on SiT-B/2.} At 400K training steps, horizontal references show published REPA (\result{PublishedREPAB2fid}; \citealp{reg}) and local REG (\result{REGbasefid}). LoopREG ($\ktr=3$) reaches FID \result{XthreeBestfid} at $\kinfer=5$ (star). Figure~\ref{fig:horizon} gives full sweeps. Local evaluations: EMA, 50K samples, 250 denoiser calls, CFG=1.}
\label{fig:main}
\end{figure}

\section{Introduction}
Diffusion Transformers such as DiT and SiT devote a fixed stack of blocks to each denoiser evaluation \citep{dit,sit}. Increasing the number of sampling steps adds computation across diffusion time. Shared recurrence offers a complementary possibility: spend more computation refining the hidden state within each evaluation, while reusing the same parameters. For this to improve generation, a learned block must remain useful when applied beyond its training horizon.

\emph{What makes recurrent computation reusable beyond its training horizon?} Our experiments identify representation supervision as an enabling ingredient. Naively repeating a block in plain SiT yields at most a marginal matched-depth improvement and degrades sharply with an extra inference visit. The same recurrent construction becomes useful under REPA alignment and substantially more effective under REG's representation entanglement \citep{repa,reg}. This contrast connects the value of extra computation to how the recurrent state is supervised.

\begin{figure}[t]
\centering\includegraphics[width=\linewidth]{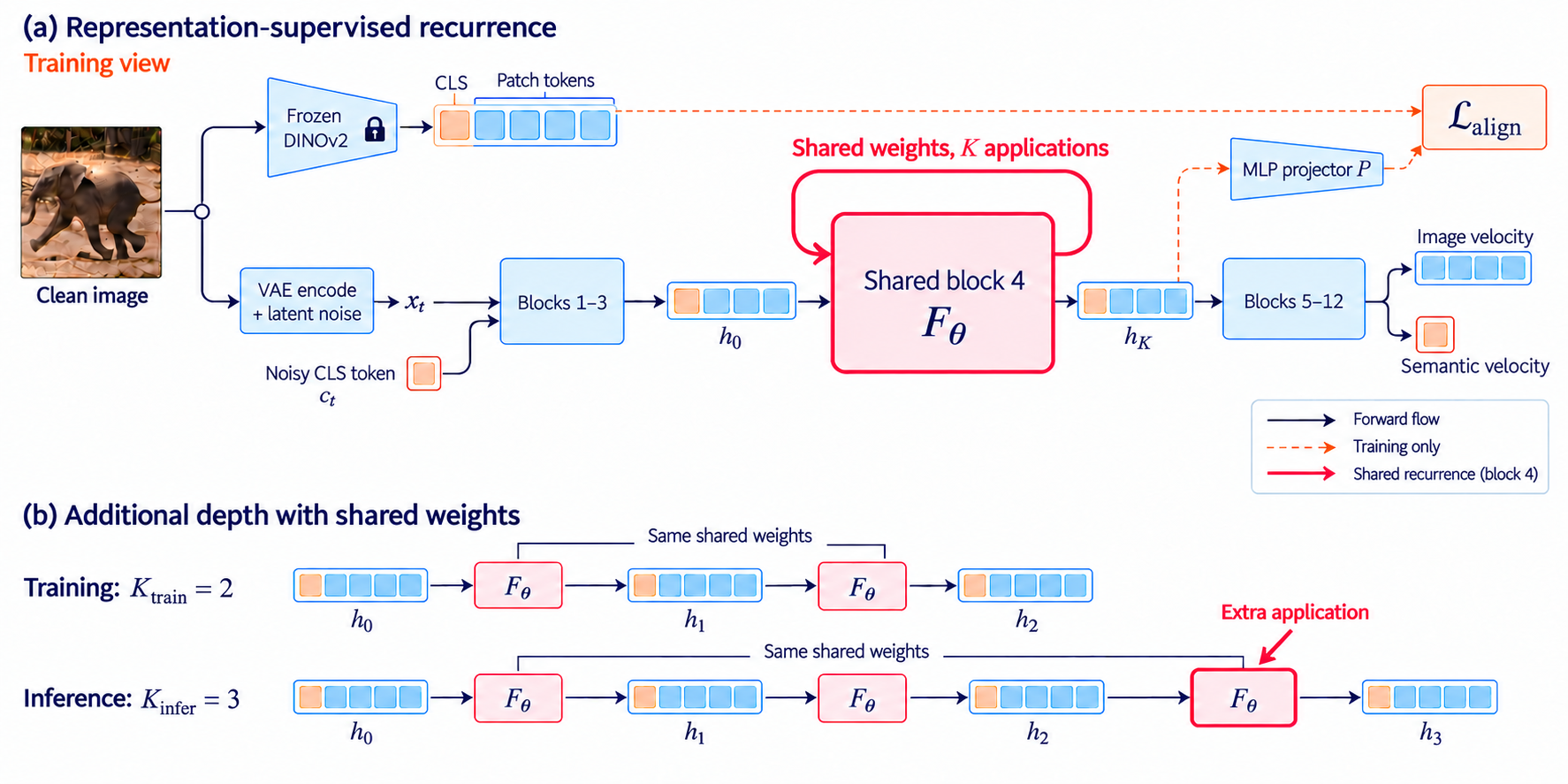}
\caption{\textbf{LoopREG: representation-supervised shared recurrence.} SiT-B/2 is shown. (a) One clean image supplies both the VAE latent, noised after encoding, and frozen DINOv2 targets. Orange denotes CLS/semantic tokens; blue denotes image/patch tokens. Alignment reads the final recurrent state through a projector. (b) Shared weights are reused beyond the training horizon, illustrated by $\ktr=2$ and $\kinfer=3$. Sampling starts from noise.}
\label{fig:architecture}
\end{figure}

We introduce \textbf{LoopREG}, a recurrent diffusion Transformer built on REG. LoopREG reuses one Transformer block at the representation-supervised state, applies alignment after the final training-time visit, and extends that block's trajectory at inference (Figure~\ref{fig:architecture}). It retains REG's joint image and semantic-token denoising objective. Inference adds visits with unchanged weights and a fixed diffusion sampling schedule.

The resulting generation gains are substantial. On ImageNet $256\times256$ at 400K training steps, LoopREG-B/2 reaches FID \result{XthreeBestfid}, compared with \result{REGbasefid} for the local REG baseline: a \result{BaselineToX3Gain}\% reduction (Figure~\ref{fig:main}). Within a LoopREG checkpoint trained with two visits, a third visit reduces FID from \result{REGtwofid} to \result{REGthreefid}. Useful extra depth also persists on SiT-L/2 and SiT-XL/2. On XL/2, extra inference depth reaches the matched-depth 400K quality at 220K training steps.

The key design choice is \emph{where} to recur. With supervision fixed at block four, extra recurrence helps at the supervised state and fails at the other tested sites. Moving supervision and recurrence together restores the benefit at other depths. A compact analysis relates this placement to direct supervision of the recurrent endpoint and interprets the finite useful inference horizon.

Our main contributions are:
\begin{enumerate}
\item \textbf{LoopREG.} We introduce representation-supervised shared recurrence for diffusion Transformers, enabling additional inference depth with the same trained parameters.
\item \textbf{Substantial generation gains.} LoopREG lowers SiT-B/2 FID from \result{REGbasefid} to \result{XthreeBestfid}, with useful recurrent-depth extrapolation across B/2, L/2, and XL/2. Training-horizon sweeps characterize the resulting quality--depth curves.
\item \textbf{A supervision-based design principle.} Plain SiT/REPA/REG controls and site/depth interventions show that representation supervision and its location determine whether extra recurrent computation is useful in the tested settings.
\end{enumerate}

\section{Related Work}
\paragraph{Diffusion and representation supervision.}
DiT replaces the denoising backbone with a Transformer, and SiT studies flow and diffusion within a stochastic-interpolant framework \citep{dit,sit}. REPA aligns projected intermediate states with clean pretrained visual representations \citep{repa}. REG jointly denoises image latents and a semantic token while retaining representation alignment \citep{reg}. SRA instead uses the model's own EMA representations for alignment \citep{sra}. We study a further capability of representation supervision: enabling a recurrent operator to improve generation beyond its training horizon. The supervision-site interventions identify where this capability emerges.

\paragraph{Looped Transformers and latent computation.}
Universal Transformers combine shared depth with adaptive computation \citep{universal}. Theoretical studies show that looped Transformers can implement programs \citep{giannou}, learn iterative data-fitting procedures \citep{yang}, and realize multistep gradient descent in controlled in-context-learning settings \citep{gatmiry}. Recurrent-depth language models explore scaling latent computation beyond a fixed forward pass \citep{recurrent}. We investigate when repeated computation improves a diffusion denoiser, using representation supervision as the central intervention.

\paragraph{Elastic depth and visual generation.}
LoopFormer conditions recurrent updates on time and step size and trains shortcut-consistent trajectories \citep{loopformer}; PonderNet learns a stopping distribution \citep{ponder}. ELT explores looped visual generation with intra-loop self-distillation and reports gains beyond training depth \citep{elt}. We instead investigate external representation supervision and its location, connecting recurrent-depth extrapolation to supervised-state dynamics. Our RandomLoop control samples terminal horizons from finite uniform distributions; heavy-tailed random horizons are also used in recurrent-depth language-model training \citep{recurrent}.

\paragraph{Dynamics and inference-time reuse.}
Deep equilibrium models learn implicitly defined fixed points \citep{deq}. Recurrent-dynamics work distinguishes regimes in which extra iterations help, preserve, or degrade predictions \citep{shallow}; training-free looping uses damped updates to mitigate failures of naive reapplication \citep{trainingfree}. LoopREG trains the shared operator explicitly and reuses its learned update at inference. Internal Guidance combines intermediate and final denoising predictions to guide sampling \citep{internalguidance}. LoopREG varies the recurrent depth of a representation-supervised block within each denoiser call. Its generation benefit depends on the supervision location and the inference horizon.

\section{LoopREG}
\label{sec:method}
LoopREG adds shared recurrence to a representation-supervised diffusion Transformer. Its defining choice is to repeat the block whose output receives representation supervision, so that the supervised state is also the state refined by extra inference computation.

\subsection{Representation-Supervised Shared Recurrence}
\paragraph{Image and representation inputs.}
Let $x_0$ be the VAE latent of a clean training image, $y$ its class label, and $z=(z_{\rm cls},z_{\rm patch})$ the frozen DINOv2 features of the same image. Following REG, we form an image state and a semantic-token state at a common time $t\sim\mathcal U[0,1]$:
\begin{equation}
 x_t=(1-t)x_0+t\epsilon,\qquad
 c_t=(1-t)z_{\rm cls}+t\epsilon_c,
 \label{eq:inputs}
\end{equation}
where $\epsilon$ and $\epsilon_c$ are independent Gaussian noise. The Transformer processes image tokens and the semantic token together, conditioned on $t$ and $y$, and predicts both velocities. Figure~\ref{fig:architecture} shows how the clean image supplies the latent and the alignment targets.

\paragraph{Reusing the supervised block.}
Write the blocks before the recurrent site as $H$, the shared block as $F_\theta$, and the remaining blocks and output projection as $G_\phi$. At a fixed sampling state $u_t=(x_t,c_t)$,
\begin{equation}
 h_0=H(u_t;t,y),\quad h_{k+1}=F_\theta(h_k;t,y),\quad
 (\hat v_K^x,\hat v_K^c)=G_\phi(h_K;t,y).
 \label{eq:dynamics}
\end{equation}
The depth $K$ counts all applications of the shared block, including its ordinary first visit. Thus $K=1$ executes it once and $K=3$ executes the same module three times. Each visit receives the preceding visit's state with the same time and class conditioning. LoopREG uses no recurrent-index embedding or per-visit parameters.

The supervision depth $d$ counts Transformer blocks in the original backbone. We co-locate the recurrent block at $\ell=d$ and read the representation after its \emph{final} recurrent visit. A nonlinear projector $P$ maps that endpoint to the clean teacher features. Alignment is evaluated once per forward pass, so the training objective supervises the endpoint of the learned trajectory. The main B/2 model uses $d=\ell=4$, and L/2 and XL/2 use $d=\ell=8$. Block labels here are one-based.

\subsection{Training and Inference}
\paragraph{Training objective.}
Let $M$ count teacher tokens, including the CLS token, and let bars denote unit normalization along the feature dimension. LoopREG retains REG's negative-cosine alignment and joint velocity losses:
\begin{align}
 \Lrep(h_K,z)&=-\frac1M\sum_{i=1}^{M}
 \big\langle\overline{P(h_K)_i},\bar z_i\big\rangle,\label{eq:alignment}\\
 \mathcal L&=\mathbb E\!\left[
 \operatorname{MSE}(\hat v_K^x,\epsilon-x_0)
 +0.03\operatorname{MSE}(\hat v_K^c,\epsilon_c-z_{\rm cls})
 +0.5\Lrep(h_K,z)\right].\label{eq:objective}
\end{align}
MSE averages over the coordinates of each prediction; the expectation averages over training examples, noise, and time. Fixed-depth training uses $K=\ktr$ throughout. Since every visit shares $F_\theta$, gradients update one operator over the states induced by its recurrent trajectory. RandomLoop instead samples a terminal horizon per optimization step, shared across distributed ranks, to train multiple usable endpoints.

\paragraph{Extending depth at inference.}
Generation initializes both $x$ and $c$ from noise. Each denoiser call evaluates Equation~\eqref{eq:dynamics} at a chosen $K=\kinfer$, and the sampler updates both states using the predicted velocities. The frozen visual teacher and clean representations are used only during training. Changing $\kinfer$ changes the number of shared-block applications within a call, while preserving all learned parameters and the diffusion time grid. For example, a model trained with $\ktr=2$ can be evaluated with $\kinfer=3$ to add one recurrent refinement at every sampling step.

\paragraph{Computation.}
For a backbone with $D$ blocks and a recurrent core of $q$ blocks, the number of block executions per denoiser call is
\begin{equation}B(K)=D+(K-1)q.\label{eq:compute}\end{equation}
The main model uses $q=1$, so one extra visit adds one block execution. For B/2, the trained $\ktr=2$ call executes 13 blocks and $\kinfer=3$ executes 14. Shared recurrence adds computation without adding copies of the block's weights. We report block executions alongside generation metrics; measured runtime also depends on the sampler and implementation.

\paragraph{Controlled variants.}
Plain LoopSiT uses the same recurrence with image-velocity regression alone. LoopREPA adds patch-token alignment with coefficient $0.5$ and has no REG semantic-token path or semantic loss. Independent-copy controls replace the shared operator with separately trained blocks. These variants distinguish the effects of representation supervision, REG's joint representation objective, and parameter reuse.

\subsection{Interpreting Supervised Recurrence}
\label{sec:theory}
\paragraph{Co-location directly supervises the recurrent endpoint.}
If the loop is before supervision, let $T_{\ell\to d}$ map its endpoint to the supervised layer. The alignment gradient is
\begin{equation}
 \nabla_{h_K}\Lrep=J_T(h_K)^\top
 \nabla_{T(h_K)}\ell_{\rm rep}(P(T(h_K)),z).
 \label{eq:locality-main}
\end{equation}
At co-location, $T=I$: the representation objective reads the recurrent endpoint directly. Earlier loops receive that gradient through intervening blocks, while a later block with distinct parameters lies outside the alignment branch. This structure provides a reason to co-locate recurrence and supervision. Section~\ref{sec:mechanism} tests the design by moving the two sites independently and jointly.

\paragraph{A shared operator learns from its own recurrent trajectory.}
Supervising the endpoint also trains the transitions that reach it. Let $g_k=\nabla_{h_k}\mathcal L$ and let $\theta$ contain the shared block's parameters. Backpropagation through $K=\ktr$ visits gives
\begin{equation}
 g_k=J_hF_\theta(h_k)^\top g_{k+1},\qquad
 \nabla_\theta\mathcal L=\sum_{k=0}^{K-1}
 J_\theta F_\theta(h_k)^\top g_{k+1}.
 \label{eq:shared-gradient}
\end{equation}
Time and class conditioning are held fixed within this recurrence. The same parameters receive updates at each visited state, including states produced by their own earlier applications. Co-location makes representation alignment part of the endpoint signal $g_K$. These two choices work together: alignment shapes the state at which recurrence ends, while sharing trains one transition over the path leading there. An unshared stack assigns these transitions to different parameter sets. This explains the design rationale for a reusable refinement operator, with its useful test-time horizon determined by the measured trajectory.

\paragraph{Useful refinement can end before equilibrium.}
Additional visits trace a longer hidden-state trajectory. Even a stable trajectory can pass through a task-preferred state before its fixed point. For the recurrence $h_{k+1}=h_\infty+\lambda(h_k-h_\infty)$, $0<\lambda<1$, and a preferred state $h_g=h_\infty+r(h_0-h_\infty)$, $0<r<1$,
\begin{equation}
 \|h_k-h_g\|^2=\|h_0-h_\infty\|^2(\lambda^k-r)^2.
 \label{eq:transient-main}
\end{equation}
The error first decreases and then increases as the trajectory passes $h_g$. This finite-horizon interpretation is consistent with LoopREG's observed FID minima beyond training and its later degradation. Appendix~\ref{app:proofs} gives the assumptions, formal statements, and derivations. We select inference depths from the measured generation curves.

\section{Experiments}
\label{sec:results}
\paragraph{Protocol.}
We evaluate class-conditional ImageNet at $256\times256$, using checkpoints trained for 400K optimization steps with global batch 256. Main sweeps use EMA weights, 50K generated samples, 250 denoiser calls with an Euler--Maruyama SDE sampler, and CFG=1. We use the local ImageNet reference and OpenAI Inception evaluator. Appendix~\ref{app:setup} gives the full sampling settings and provenance.

\begin{table}[!t]
\centering\small\setlength{\tabcolsep}{3.2pt}
\begin{tabular}{llrrrrrrr}
\toprule
Model & Steps & $K_{\mathrm{train}}$ & $K_{\mathrm{infer}}$ & FID $\downarrow$ & sFID $\downarrow$ & IS $\uparrow$ & P $\uparrow$ & R $\uparrow$ \\
\midrule
\rowcolor{black!4}SiT-B/2 (local) & 400K & 1 & 1 & \result{SITbasefid} & \result{SITbasesfid} & \result{SITbaseis} & \result{SITbaseprecision} & \result{SITbaserecall} \\
\rowcolor{black!4}REPA-B/2\textsuperscript{*} & 400K & -- & -- & \result{PublishedREPAB2fid} & \result{PublishedREPAB2sfid} & \result{PublishedREPAB2is} & \result{PublishedREPAB2precision} & \result{PublishedREPAB2recall} \\
\rowcolor{black!4}REG-B/2 (local) & 400K & 1 & 1 & \result{REGbasefid} & \result{REGbasesfid} & \result{REGbaseis} & \result{REGbaseprecision} & \result{REGbaserecall} \\
\addlinespace[3pt]
LoopSiT-B/2 & 400K & 2 & 2 & \textbf{\result{SiTtwofid}} & \textbf{\result{SiTtwosfid}} & \textbf{\result{SiTtwois}} & \textbf{\result{SiTtwoprecision}} & \textbf{\result{SiTtworecall}} \\
LoopSiT-B/2 & 400K & 2 & 3 & \result{SiTthreefid} & \result{SiTthreesfid} & \result{SiTthreeis} & \result{SiTthreeprecision} & \result{SiTthreerecall} \\
\addlinespace[3pt]
LoopREPA-B/2 & 400K & 2 & 2 & \result{REPAtwofid} & \textbf{\result{REPAtwosfid}} & \result{REPAtwois} & \result{REPAtwoprecision} & \textbf{\result{REPAtworecall}} \\
LoopREPA-B/2 & 400K & 2 & 3 & \textbf{\result{REPAthreefid}} & \result{REPAthreesfid} & \textbf{\result{REPAthreeis}} & \textbf{\result{REPAthreeprecision}} & \result{REPAthreerecall} \\
\addlinespace[3pt]
LoopREG-B/2 & 400K & 2 & 2 & \result{REGtwofid} & \result{REGtwosfid} & \result{REGtwois} & \result{REGtwoprecision} & \textbf{\result{REGtworecall}} \\
LoopREG-B/2 & 400K & 2 & 3 & \textbf{\result{REGthreefid}} & \textbf{\result{REGthreesfid}} & \textbf{\result{REGthreeis}} & \textbf{\result{REGthreeprecision}} & \result{REGthreerecall} \\
LoopREG-B/2 & 400K & 3 & 3 & \result{XthreeMatchedfid} & \result{XthreeMatchedsfid} & \result{XthreeMatchedis} & \result{XthreeMatchedprecision} & \textbf{\result{XthreeMatchedrecall}} \\
LoopREG-B/2 & 400K & 3 & 5 & \textbf{\result{XthreeBestfid}} & \textbf{\result{XthreeBestsfid}} & \textbf{\result{XthreeBestis}} & \textbf{\result{XthreeBestprecision}} & \result{XthreeBestrecall} \\
\midrule
\rowcolor{black!4}SiT-L/2\textsuperscript{*} & 400K & -- & -- & \result{PublishedSiTL2fid} & \result{PublishedSiTL2sfid} & \result{PublishedSiTL2is} & \result{PublishedSiTL2precision} & \result{PublishedSiTL2recall} \\
\rowcolor{black!4}REPA-L/2\textsuperscript{*} & 400K & -- & -- & \result{PublishedREPAL2fid} & \result{PublishedREPAL2sfid} & \result{PublishedREPAL2is} & \result{PublishedREPAL2precision} & \result{PublishedREPAL2recall} \\
\rowcolor{black!4}REG-L/2\textsuperscript{*} & 400K & -- & -- & \result{PublishedREGL2fid} & \result{PublishedREGL2sfid} & \result{PublishedREGL2is} & \result{PublishedREGL2precision} & \result{PublishedREGL2recall} \\
\addlinespace[3pt]
LoopREG-L/2 & 400K & 2 & 2 & \result{Ltwofid} & \result{Ltwosfid} & \result{Ltwois} & \result{Ltwoprecision} & \textbf{\result{Ltworecall}} \\
LoopREG-L/2 & 400K & 2 & 3 & \textbf{\result{Lthreefid}} & \textbf{\result{Lthreesfid}} & \textbf{\result{Lthreeis}} & \textbf{\result{Lthreeprecision}} & \result{Lthreerecall} \\
\midrule
\rowcolor{black!4}SiT-XL/2\textsuperscript{*} & 7M & -- & -- & \result{PublishedSiTXL2fid} & \result{PublishedSiTXL2sfid} & \result{PublishedSiTXL2is} & \result{PublishedSiTXL2precision} & \result{PublishedSiTXL2recall} \\
\rowcolor{black!4}REPA-XL/2\textsuperscript{*} & 4M & -- & -- & \result{PublishedREPAXL2fid} & \result{PublishedREPAXL2sfid} & \result{PublishedREPAXL2is} & \result{PublishedREPAXL2precision} & \result{PublishedREPAXL2recall} \\
\rowcolor{black!4}REG-XL/2\textsuperscript{*} & 400K & -- & -- & \result{PublishedREGXL2fid} & \result{PublishedREGXL2sfid} & \result{PublishedREGXL2is} & \result{PublishedREGXL2precision} & \result{PublishedREGXL2recall} \\
\addlinespace[3pt]
LoopREG-XL/2 & 400K & 2 & 2 & \result{MainXLtwofid} & \result{MainXLtwosfid} & \result{MainXLtwois} & \result{MainXLtwoprecision} & \textbf{\result{MainXLtworecall}} \\
LoopREG-XL/2 & 400K & 2 & 3 & \textbf{\result{MainXLthreefid}} & \textbf{\result{MainXLthreesfid}} & \textbf{\result{MainXLthreeis}} & \textbf{\result{MainXLthreeprecision}} & \result{MainXLthreerecall} \\
\bottomrule
\end{tabular}

\caption{\textbf{Generation quality across model scales.} Horizontal rules separate backbones; spacing separates method families. Shaded rows are baselines; an asterisk denotes published results from \citet{reg}, Table~9 (SiT/REPA attributed there to \citet{sit,repa}). Training steps are shown explicitly; the published REPA-XL/2 reference uses 4M steps. All other rows are local evaluations. Bold marks the best displayed metric within each multi-row checkpoint group; best depths are selected within the reported sweeps. P/R: precision/recall. Appendix~\ref{app:setup} gives provenance, including the L/2 server summary.}
\label{tab:main}
\end{table}

\subsection{Generation Gains from Test-Time Depth}
\begin{figure}[t]
\centering\includegraphics[width=.94\linewidth]{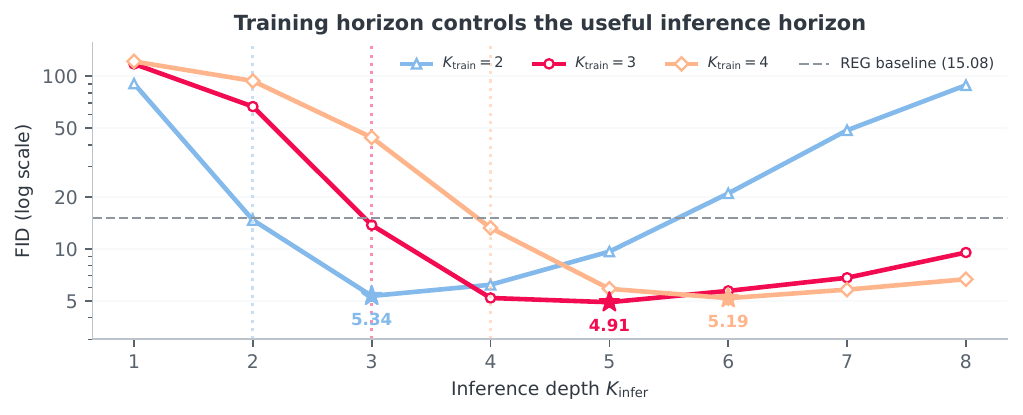}
\caption{\textbf{Training horizon shifts the useful inference horizon.} Dotted lines mark training horizons; stars and labels mark the best observed FID. The dashed horizontal line shows the local REG baseline. All three training horizons produce large improvements over this baseline, with optima moving to later recurrent depths. The logarithmic vertical axis preserves the severe under-roll failures.}
\label{fig:horizon}
\end{figure}

\paragraph{LoopREG delivers large gains over REG and its trained endpoint.}
For LoopREG-B/2 trained at $\ktr=2$, FID improves from \result{REGtwofid} to \result{REGthreefid} and remains below the matched value at $\kinfer=4$. The relative reduction at $\kinfer=3$ is \result{REGGain}\%. Increasing the training horizon shifts the minimum: $\ktr=3$ reaches FID \result{XthreeBestfid} at $\kinfer=5$, while $\ktr=4$ reaches \result{Train4K6FID} at $\kinfer=6$. Relative to the local REG baseline at \result{REGbasefid}, the best result reduces FID by \result{BaselineToX3Gain}\%. Figure~\ref{fig:horizon} shows how all three recurrent horizons cross below the baseline before reaching their finite optima. Within-checkpoint sweeps use unchanged weights. Across training horizons, equal optimization steps involve different training compute, as quantified by Equation~\eqref{eq:compute}.

\paragraph{Representation supervision unlocks recurrent extrapolation.}
Plain LoopSiT at matched depth slightly improves on the local SiT baseline (FID \result{SITbasefid} to \result{SiTtwofid}), but an additional visit worsens FID to \result{SiTthreefid}. LoopREPA improves by \result{REPAGain}\% from $\kinfer=2$ to $\kinfer=3$, despite having no REG semantic token, then degrades at $\kinfer=4$. Alignment alone therefore enables useful extrapolation in this control, while REG substantially increases the gain and useful recurrent horizon.

\subsection{Scaling to Larger Backbones}
\label{sec:scale}
The extrapolation benefit persists as the backbone grows. At 400K steps, LoopREG-L/2 improves from FID \result{Ltwofid} at $\kinfer=2$ to \result{Lthreefid} at $\kinfer=3$. LoopREG-XL/2 uses 28 blocks, supervision at depth eight, and shared block eight. It improves from \result{XLtwofid} to \result{XLthreefid}, a \result{XLGain}\% reduction, with sFID also improving. Table~\ref{tab:main} places these results beside the published SiT and REG baselines at each scale. Together with the B/2 results, these experiments establish useful recurrent extrapolation across three model scales.

\paragraph{One extra block application yields a large quality return.}
Table~\ref{tab:compute-return} isolates the added recurrent depth within each checkpoint. Moving from $\kinfer=2$ to $3$ adds one shared-block execution per denoiser call across all three backbones, with unchanged weights and the same 250-call sampler. The relative increase in block executions becomes smaller for deeper backbones, while each model retains a generation benefit. This gives LoopREG a direct way to spend additional computation inside the denoiser, complementary to increasing the number of sampling steps. Block counts measure executed Transformer depth; their runtime cost also depends on the backbone and implementation.

\begin{table}[!htbp]
\centering\small\setlength{\tabcolsep}{3.5pt}
\begin{tabular}{lrrrr}
\toprule
Backbone & Blocks/call & Added blocks (\%) & FID ($\kinfer$: $2\to3$) & FID reduction (\%)\\
\midrule
SiT-B/2 & 13$\to$14 & 7.69 & 14.71$\to$5.34 & \textbf{63.67}\\
SiT-L/2 & 25$\to$26 & 4.00 & 4.23$\to$3.21 & \textbf{24.13}\\
SiT-XL/2 & 29$\to$30 & 3.45 & 3.38$\to$2.84 & \textbf{16.10}\\
\bottomrule
\end{tabular}

\caption{\textbf{Return from one extra shared-block application.} All models train with $\ktr=2$ for 400K steps. Added block executions are relative to matched $\kinfer=2$; FID reductions compare the same checkpoint at $\kinfer=2$ and $3$.}
\label{tab:compute-return}
\end{table}

\begin{figure}[t]
\centering\includegraphics[width=\linewidth]{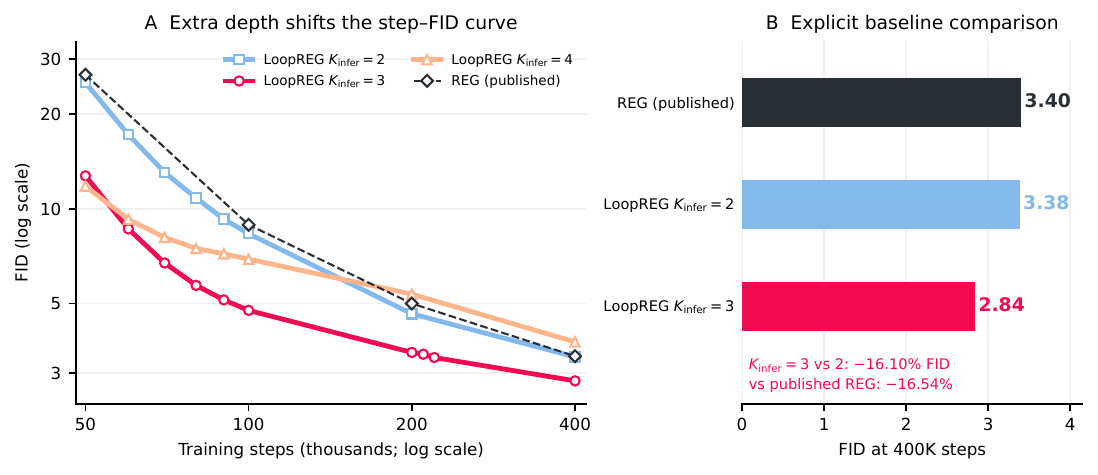}
\caption{\textbf{Extra recurrent computation improves the optimization-step--FID curve.} Left: fixed $\kinfer=2,3,4$ curves from 50K steps along one XL training run, with published REG reference points. Right: 400K results; $\kinfer=2,3$ share one checkpoint, while REG is a published reference. Lines connect evaluated checkpoints; 210K/220K include $\kinfer=3$ only.}
\label{fig:xlfrontier}
\end{figure}

\paragraph{Reaching lower FID with fewer optimization steps.}
Along the same XL training trajectory, 220K at $\kinfer=3$ reaches FID \result{XLearlyfid}, compared with \result{XLtwofid} at 400K and $\kinfer=2$: \result{XLLocalStepsSaved}\% fewer optimization steps at a slightly lower FID. Against published REG-XL thresholds \citep{reg}, the earliest evaluated crossings use 40K/60K/100K/220K steps versus 50K/100K/200K/400K (Figure~\ref{fig:xlfrontier}). These are optimization-step comparisons with additional inference recurrence; measured runtime savings require a separate assessment. Appendix~\ref{app:xl} reports the full checkpoint grid and metric comparisons.

\paragraph{Fidelity and coverage.}
The 400K XL $\kinfer=2$ to $3$ improvement accompanies higher precision and lower recall (\result{XLtworecall} to \result{XLthreerecall}). For the 220K $\kinfer=3$ versus 400K $\kinfer=2$ comparison, sFID improves from \result{XLtwosfid} to \result{XLearlysfid}, while recall falls to \result{XLearlyrecall}. At 60K--100K, $\kinfer=3$ improves FID but slightly worsens sFID. Precision and recall in Table~\ref{tab:main} make this fidelity--coverage change explicit. Training with random horizons offers an alternative when retaining coverage is a priority (Section~\ref{sec:policies}).

\subsection{Where Should LoopREG Recur?}
\label{sec:mechanism}
\begin{figure}[t]
\centering\includegraphics[width=\linewidth]{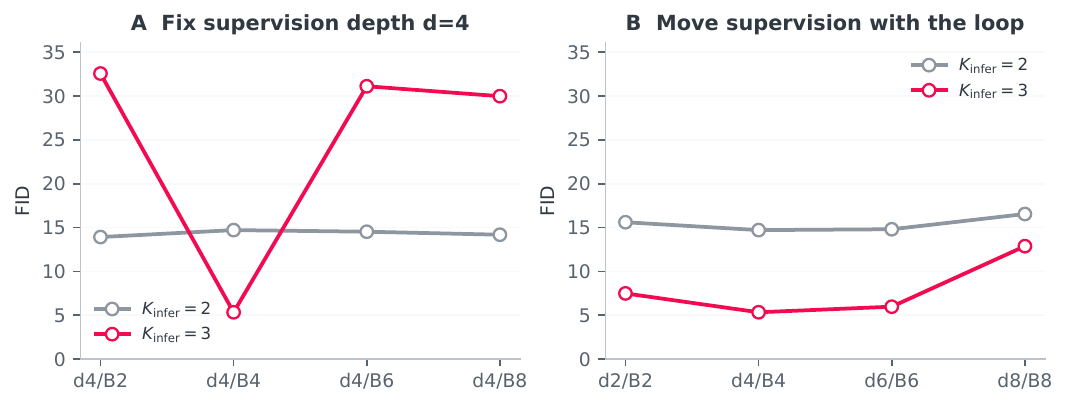}
\caption{\textbf{Useful recurrence follows representation supervision in the tested interventions.} Left: fixed supervision at depth four, varied loop block. Right: supervision and recurrence moved together. Both panels reuse the d4/B4 anchor. Each model trains at $\ktr=2$.}
\label{fig:site}
\end{figure}

\paragraph{Useful extrapolation depends on the supervised state.}
We first hold REG supervision at $d=4$ and move the recurrent block. At matched $\kinfer=2$, all four tested sites have similar FID, but the unseen $\kinfer=3$ response separates them sharply (Figure~\ref{fig:site} and Table~\ref{tab:site}). B4 improves, while B2, B6, and B8 degrade. Thus matched-depth quality alone does not predict useful extrapolation.

We then move supervision with the loop. Co-located d2/B2, d6/B6, and d8/B8 each improve at $\kinfer=3$, with a weaker effect at depth eight. The benefit moves with supervision, supporting the co-located design in Section~\ref{sec:theory}. The two groups share the same d4/B4 anchor.

\paragraph{Representation geometry tracks useful recurrent refinement.}
We probe the recurrent states of the two LoopREG-B/2 checkpoints on 1,024 forward-noised ImageNet images at five diffusion times. Patch CKA compares mean hidden patch features with mean DINOv2 patch features, using a fixed 256-image subset. Spatial correlation compares pairwise patch-token similarities, using 64 sampled positions across all 1,024 images. These complementary measurements describe semantic organization and spatial structure along the recurrent trajectory.

\begin{figure}[!htbp]
\centering\includegraphics[width=\linewidth]{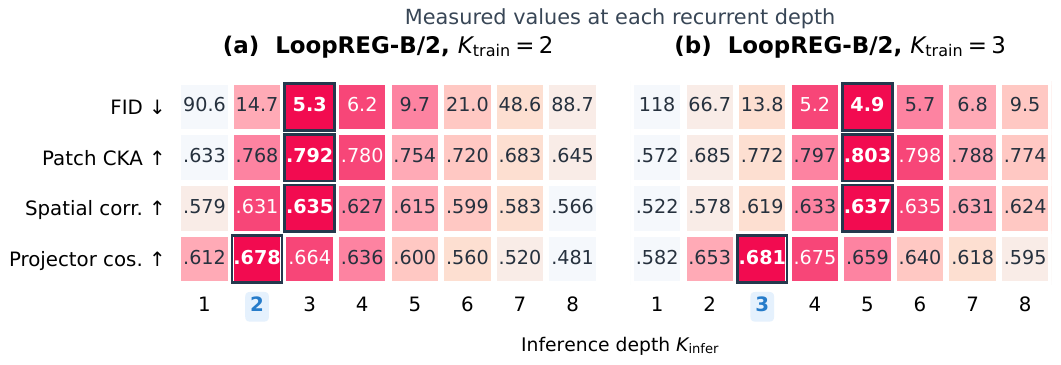}
\caption{\textbf{Useful extra depth improves representation geometry.} Cells show measured values; darker shading indicates a better rank \emph{within each row}, and outlines mark that row's optimum. Blue depth labels mark $\ktr$. Patch CKA, spatial correlation, and FID prefer the same extra depth for each LoopREG checkpoint. Native projector cosine instead peaks at the trained depth. Representation measurements use the final N=1,024 probe (CKA subset 256), averaged over five diffusion times; FID uses the separate 50K generation evaluation.}
\label{fig:refinement}
\end{figure}

Figure~\ref{fig:refinement} displays every tested depth together with generation FID. For $\ktr=2$, patch CKA and spatial correlation both peak at $\kinfer=\result{Refine2PatchCKABestK}$, where FID is lowest. Training at $\ktr=3$ moves these geometry peaks to $\kinfer=\result{Refine3PatchCKABestK}$, again matching the best generation depth. The extended trajectory therefore contains states with stronger teacher-referenced geometry than the state at which training ends. This correspondence connects the generation gains to refinement of the representation selected by LoopREG's supervision.

\paragraph{The trained projection and extended trajectory play different roles.}
Native patch-projector cosine reaches its maximum at the trained depth in both checkpoints. The projector thus favors the supervised endpoint, while the best generation states occur farther along the trajectory. This distinction fits the finite-horizon interpretation in Section~\ref{sec:theory}: a block can continue to improve task-relevant geometry after reaching the endpoint used by its training objective. Appendix~\ref{app:dynamics} provides all five models and individual diffusion times, including the plain-recurrence control. Paired output-depth interventions in Appendix~\ref{app:causal} further identify a contribution from REG's semantic-state updates.

\subsection{Choosing a Depth Policy}
\label{sec:policies}
Fixed training horizons give LoopREG its strongest peak FID. They also concentrate generation around a narrower depth range: at its best B/2 depth, the $\ktr=3$ model has precision \result{XthreeBestprecision} and recall \result{XthreeBestrecall}, versus the REG baseline's \result{REGbaseprecision}/\result{REGbaserecall}. RandomLoop broadens the useful range and preserves more coverage. Training with $K\in\{1,2,3\}$ yields a best observed FID of \result{Random123fid} with recall \result{Random123recall}. This offers a choice between peak fidelity and flexibility in the number of inference visits.

A timestep-dependent policy adds another operating point. Applying $\kinfer=3$ for $t\geq0.6$ and $\kinfer=2$ for $t<0.6$ reaches FID \result{Adaptivefid}, between the two fixed-depth results, with recall \result{Adaptiverecall}. It reduces total block executions by \result{AdaptiveBlockSavings}\% relative to fixed $\kinfer=3$. Appendix~\ref{app:controls} reports the complete random-depth curves, compute accounting, and unshared controls.

\section{Conclusion}
LoopREG makes additional test-time depth useful by repeatedly applying a shared Transformer block at a representation-supervised state. It reduces SiT-B/2 FID from \result{REGbasefid} to \result{XthreeBestfid} at 400K training steps and extends recurrent-depth gains to L/2 and XL/2. The supervision-site interventions identify a transferable design choice within these models: useful recurrence follows the supervised representation. These results establish a practical way to improve diffusion generation by extending learned recurrent computation.

\paragraph{Scope and future directions.}
The experiments cover class-conditional ImageNet, SiT backbones, and DINOv2 supervision. Useful depth has a finite optimum, and improved FID often accompanies lower recall. Extending the approach to other generation settings and selecting depths without a generation sweep are natural next steps.
\label{sec:mainend}

\section*{Reproducibility Statement}
Appendix~\ref{app:setup} specifies training, sampling, evaluation, and pretrained assets. Appendices~\ref{app:sweeps} and~\ref{app:dynamics} report final-checkpoint generation sweeps and representation trajectories. The explanatory derivations, paired output-depth interventions, and depth-selection diagnostics appear in Appendices~\ref{app:proofs}, \ref{app:causal}, and~\ref{app:depthselection}. All reported numerical results are drawn from recorded experimental evaluations.
\section*{AI Assistance Disclosure}
Generative AI tools assisted with manuscript drafting and revision, literature retrieval, code auditing, result interpretation, development of the dynamical-systems framework and proofs, experiment planning, architecture-figure generation and editing, and implementation of data-processing, visualization, and evaluation scripts. Numerical results in this manuscript are drawn from the project's recorded experimental evidence.
\bibliography{refs}
\bibliographystyle{iclr2027_conference}
\clearpage
\appendix
\raggedbottom
\section{Experimental Setup and Evidence Provenance}
\label{app:setup}
We study class-conditional ImageNet generation at $256\times256$. Generation metrics and representation probes are retained in separate tables with their checkpoint, recurrent-depth, and evaluation settings. All numerical tables and inline results are generated from these records. The final 1,024-image DINO probe supplies the representation analysis; output-depth interventions and smaller-sample depth-selection diagnostics use their own evaluation protocols.

\paragraph{Training.}
SiT-B/2 has 12 blocks of width 768 with 12 heads; SiT-L/2 has 24 blocks of width 1024 with 16 heads. Image latents have four channels and spatial size $32\times32$, with latent patch size two. The semantic teacher is DINOv2 ViT-B/14 \citep{dino}. Training uses global batch 256, AdamW with learning rate $10^{-4}$, betas $(0.9,0.999)$, epsilon $10^{-8}$, no weight decay, gradient clipping at one, class dropout $0.1$, FP16 mixed precision, and TF32. The default launcher uses eight GPUs and checkpoints every 10K steps. RandomLoop's horizon draw is a deterministic function of the experiment seed and optimization step and is synchronized across ranks; equal configured weights implement a uniform distribution over the specified choices. Horizon sampling has no learned halting mechanism. Nominally equal step counts do not match training FLOPs across horizons or supervision methods.

\paragraph{Sampling.}
The sampler constructs 250 time points from $1$ to $0.04$ and appends zero. It applies 249 stochastic Euler--Maruyama updates and a final deterministic drift update, totaling 250 network calls at CFG=1. REG jointly evolves image and semantic-token states. Network inputs/weights are FP32 with TF32 enabled; velocities and state updates are promoted to FP64. The VAE decoder is evaluated separately. Generated images are clamped, multiplied by 255, and converted to uint8 by truncation. The evaluator reads a local ImageNet reference and the local OpenAI TensorFlow Inception graph. FID, sFID, IS, precision, and recall are imported as reported, without recomputing them from unavailable image batches.

The configured global seed is zero and the rank seed is $\mathrm{seed}\times\mathrm{world\_size}+\mathrm{rank}$. Raw artifacts show different world sizes or per-GPU batches across runs; therefore these are not matched-noise paired comparisons. The source's resume branch skips the sampler for completed batches and does not advance the random-number stream through their internal SDE noise draws. This prevents a blanket claim of bit-identical resumed and uninterrupted generation. The available artifacts do not establish which final cells, if any, were affected. The paired output-depth experiments below instead use fresh full runs and record their complete random-number traces.

\paragraph{Late evidence and precision.}
The final N=1,024 mechanism bundle and unshared-x3 K3/K6 metrics supersede earlier development results or missing-data notes. The representation analysis in Appendix~\ref{app:dynamics} uses this final bundle. The separate velocity probe remains N=256 because no larger velocity probe is supplied. The L/2 main-result curve is supplied as the experiment owner's final server-summary Markdown, at four decimal places for FID; it is parsed automatically, but its original metric/configuration JSON is not yet part of this evidence set. The verified launcher and snapshot checkpoint-status metadata are corroborating implementation/status evidence, not a replacement for those metric artifacts.

\paragraph{Baseline and selection scope.}
Local REG-B/2 is the comparison baseline; proximity to the published REG result is a sanity check, not proof of evaluator identity. No independently evaluated local non-loop REPA baseline or local non-loop L/2 baseline is inserted. Best inference depths are descriptive minima over observed sweeps. Appendix~\ref{app:sweeps} presents full inference-depth sweeps at the final core-model checkpoints; the selected configurations include all evaluated inference depths, including unfavorable outcomes. No standard errors or confidence intervals are synthesized from a single seed.

\paragraph{Independent output-depth evaluations.}
Appendix~\ref{app:causal} adds paired 50K evaluations at generation seeds 2 and 3 for both train-two and train-three B/2 checkpoints. GPU/batch configuration is fixed within each checkpoint study. Fresh full runs reproduce the matched and one-step-extrapolated endpoints and evaluate mixed output depths. The mixed policies require two network evaluations per state update. They are distinct records from the historical depth sweeps and do not resolve missing provenance for other historical checkpoints.

\paragraph{Pretrained assets.}
ImageNet provides the training images and class labels. Image latents use the pretrained VAE, and alignment targets use frozen DINOv2 ViT-B/14 features. The visual teacher is used during training and representation analysis; generation requires the trained diffusion model and VAE decoder.

\section{Dynamical Interpretation and Derivations}
\label{app:proofs}
LoopREG has two interacting iteration scales: recurrent applications evolve the hidden state within a denoiser call, while diffusion steps evolve the image and semantic states across calls. We analyze where supervision enters the first trajectory, why its useful horizon can be finite, and how its outputs affect the second trajectory. The position interventions appear in Section~\ref{sec:mechanism}; the output-path interventions appear in Appendix~\ref{app:causal}.

\subsection{Supervision Locality}
\begin{proposition}[Gradient at the recurrent endpoint]
\label{prop:locality}
Let $d$ be the supervision depth and $\ell$ the loop site. For $\ell\leq d$, write $T=T_{\ell\to d}$ for the intervening transformation and $\Lrep=\ell_{\rm rep}(P(T(h_K)),z)$. At differentiable points,
\begin{equation}
 \nabla_{h_K}\Lrep=J_T(h_K)^\top J_P(T(h_K))^\top
 \nabla_1\ell_{\rm rep}(P(T(h_K)),z).
 \label{eq:pullback}
\end{equation}
At $\ell=d$, $T=I$. If $\ell>d$ and the loop parameters are distinct from upstream parameters, this alignment loss has zero derivative with respect to the loop parameters.
\end{proposition}
Co-location makes the projector read the recurrent endpoint directly. Earlier recurrence receives alignment gradients through the intervening layers; later recurrence is trained by downstream denoising losses but lies outside the alignment branch. This gradient structure motivates the hypothesis that co-location favors useful extrapolation. Moving supervision and recurrence separately and jointly tests that hypothesis, rather than selecting a block from matched-depth quality alone.

\subsection{Finite Useful Horizons}
\begin{proposition}[Task-optimal transient state]
\label{prop:transient}
Consider $h_{k+1}=h_\infty+\lambda(h_k-h_\infty)$, with $0<\lambda<1$ and $h_0\ne h_\infty$. For a task-preferred state $h_g=h_\infty+r(h_0-h_\infty)$, $0<r<1$, the increments shrink geometrically, while $Q(k)=\|h_k-h_g\|^2$ has continuous minimum at
\begin{equation}
 k_{\rm cont}^{\star}=\frac{\log r}{\log\lambda}.
 \label{eq:optimum}
\end{equation}
An integer minimizer is one of the two bracketing integers, selected by comparing $Q$. Beyond the crossing, $Q$ increases toward a positive limit.
\end{proposition}
Stability and task quality can therefore prefer different stopping times. The proposition characterizes a stable scalar recurrence; the learned Transformer's finite FID optimum is an empirical observation. It motivates studying the whole recurrent trajectory rather than treating smaller state increments as a stopping criterion.

For a family satisfying $\lambda_{\ktr}^{\ktr}=c$ and a shared task-preferred progress $r_g$, with $0<c<1$ and $0<r_g<1$, substitution gives
\begin{equation}
 k_{\rm cont}^{\star}=\ktr\frac{\log r_g}{\log c}.
 \label{eq:horizon}
\end{equation}
The continuous optimum lies beyond training exactly when $r_g<c$; Appendix~\ref{app:proofs} derives the stricter condition for an extra integer application to help. Under the shared-progress assumptions, a longer training horizon distributes latent progress over smaller updates and moves the optimum to a later depth. The observed $2\to3$, $3\to5$, and $4\to6$ horizons are consistent with this direction. The constants specify the assumed family; selecting real-model depths requires estimating their task dependence. The derivations and a schematic follow below; Appendix~\ref{app:depthselection} evaluates candidate depth-selection models.

\subsection{Semantic Feedback across Sampling Steps}
\label{sec:feedback}
For REG, the recurrent computation returns image and semantic velocities $(\hat v_K^x,\hat v_K^c)$. Under the linear interpolant, CFG=1, and squared diffusion amplitude $g(t)^2=2t$, the implemented reverse-time drifts at $t>0$ are
\begin{equation}
 b_K^x=x+(2-t)\hat v_K^x(x,c,t,y),\qquad
 b_K^c=c+(2-t)\hat v_K^c(x,c,t,y).
 \label{eq:jointdrift}
\end{equation}
An Euler--Maruyama step updates each state with its corresponding drift and additive noise; the time increment $\Delta t_j$ is negative.

\begin{proposition}[Delayed image effect of a semantic update]
\label{prop:feedback}
Start two updates from the same $(x_j,c_j)$ with identical noise increments. Hold the image velocity fixed and change the semantic velocity by $\delta v_j^c$. Then
\begin{equation}
 \delta x_{j+1}=0,\qquad
 \delta c_{j+1}=\Delta t_j(2-t_j)\delta v_j^c.
 \label{eq:delay}
\end{equation}
If the next image update uses the same function $b^x$ in both trajectories and the same noise,
\begin{equation}
 \delta x_{j+2}=\Delta t_{j+1}\big[
 b^x(x_{j+1},c_{j+1}+\delta c_{j+1},t_{j+1},y)
 -b^x(x_{j+1},c_{j+1},t_{j+1},y)\big].
 \label{eq:feedback}
\end{equation}
Changing only the final semantic output leaves the final image unchanged.
\end{proposition}
Thus an extra-depth semantic update can affect images through subsequent evaluations of the image vector field. This identifies a pathway to intervene on: evaluate two depths at the same current joint state, and choose the image and semantic outputs separately. Our directional hypothesis is that, when the image output already uses extra depth, the extra-depth semantic output further improves generation. The structure determines the intervention; the sign of its FID effect is tested across two training horizons in Appendix~\ref{app:causal}.

\subsection{Derivations}
\paragraph{Proof of Proposition~\ref{prop:locality}.}
For $s=T(h_K)$ and $a=P(s)$, the total differential is
$d\Lrep=\nabla_1\ell_{\rm rep}(a,z)^\top J_P(s)J_T(h_K)\,dh_K$.
Transposing gives Equation~\eqref{eq:pullback}. At co-location $J_T=I$. If alignment is evaluated before the recurrent block, its computational graph has no dependency on the distinct parameters of that later block, so that partial derivative is zero. This claim presumes those parameters are not also tied to upstream modules; the audited local-block implementation satisfies this condition. Other downstream losses may still provide nonzero gradients.

For a shared core, set $J_j=\partial F_\theta(h_j)/\partial h_j$ and $A_j=\partial F_\theta(h_j)/\partial\theta$. Holding the prefix state fixed with respect to the core parameters, define adjoints $g_K=\nabla_{h_K}\mathcal L$ and $g_j=J_j^\top g_{j+1}$. Then
\begin{equation}
 \nabla_\theta\mathcal L=\sum_{j=0}^{K-1} A_j^\top g_{j+1}.
\end{equation}
This is the ordinary backpropagation-through-time identity. The shared operator aggregates gradients over states reached by its own trajectory. Contributions can reinforce or cancel; an unshared stack instead updates separate parameter sets.

\paragraph{Proof of Proposition~\ref{prop:transient}.}
Induction gives $h_k=h_\infty+\lambda^k(h_0-h_\infty)$. Consequently,
\begin{align}
 \|h_{k+1}-h_k\|&=(1-\lambda)\lambda^k\|h_0-h_\infty\|,\\
 Q(k)&=\|h_0-h_\infty\|^2(\lambda^k-r)^2.
\end{align}
For continuous $k\geq0$, $\lambda^k$ decreases strictly from one to zero and crosses $r$ once, at $\log(r)/\log(\lambda)$. Before the crossing, the error strictly decreases; afterwards it strictly increases. A discrete minimizer must be one of the bracketing integers. For $m=\lfloor k_{\rm cont}^{\star}\rfloor$, choose $m$ if
$r\geq(\lambda^m+\lambda^{m+1})/2$, and $m+1$ otherwise, with a tie at equality. Thus rounding $k_{\rm cont}^{\star}$ to the nearest integer is not generally correct because the trajectory is nonlinear in depth. The limiting error is $r^2\|h_0-h_\infty\|^2>0$.

The same derivation applies to a vector trajectory restricted to a single stable eigenmode when the preferred state lies on that trajectory. General multidimensional dynamics need not be unimodal, normal, or contractive. Neither shrinking observed increments nor a U-shaped FID curve establishes a spectral-radius bound for a Transformer.

\paragraph{Latent discretization.}
Taking logarithms of $\lambda_{\ktr}^{\ktr}=c$ yields $\log\lambda_{\ktr}=\log c/\ktr$. Substitution into the continuous optimum gives Equation~\eqref{eq:horizon}. For $r_g<c$, both logarithms are negative and $\log r_g/\log c>1$, giving a continuous optimum beyond the training horizon. The through-origin descriptive fit to the observed horizon pairs has coefficient \result{HorizonSlope}. This summarizes the observed horizon shift under the latent-progress model; estimating its variability requires additional training configurations.

\paragraph{When does one extra application improve the task error?}
Write $C=\|h_0-h_\infty\|^2>0$ and $\lambda=\lambda_{\ktr}$, so $\lambda^{\ktr}=c$. Subtracting the two task errors gives
\begin{equation}
 Q(\ktr+1)-Q(\ktr)=C c(\lambda-1)\big[c(1+\lambda)-2r_g\big].
\end{equation}
Since $c>0$ and $\lambda<1$, one extra integer application improves $Q$ if and only if
\begin{equation}
 r_g<\frac{c(1+\lambda)}{2}.
 \label{eq:extra-step-condition}
\end{equation}
Equality gives a tie. This is stricter than $r_g<c$, which only places the continuous optimum beyond training. For example, $\lambda=1/2$, $\ktr=2$ and $r_g=0.24$ place the continuous optimum after two, yet the discrete optimum remains two. The criterion identifies the missing task-dependent quantity: the desired progress $r_g$ must be estimated independently of the observed best depth before this model can select an inference horizon.

\paragraph{Proof of Proposition~\ref{prop:feedback}.}
For a state $u\in\{x,c\}$, the linear interpolant gives
\begin{equation}
 \operatorname{score}_K^u=\frac{-(1-t)\hat v_K^u-u}{t},\qquad
 b_K^u=\hat v_K^u-\tfrac12(2t)\operatorname{score}_K^u=u+(2-t)\hat v_K^u.
\end{equation}
The sampler evaluates this expression at positive times, including $t=0.04$ for its final update to zero. Its stochastic steps are
\begin{align}
 x_{j+1}&=x_j+\Delta t_j b^x(x_j,c_j,t_j,y)+\sqrt{2t_j|\Delta t_j|}\,\xi_j^x,\\
 c_{j+1}&=c_j+\Delta t_j b^c(x_j,c_j,t_j,y)+\sqrt{2t_j|\Delta t_j|}\,\xi_j^c.
\end{align}
Take two steps from the same current state with the same two noise tensors. Their image drifts are identical, giving $\delta x_{j+1}=0$. Subtracting the semantic updates cancels the noise and the current state, giving Equation~\eqref{eq:delay}. At the next step, subtracting the image updates again cancels the common image state and noise; the difference of the image drift evaluations gives Equation~\eqref{eq:feedback}. The final sampler call returns only the image update, so its semantic output cannot affect the returned image.

For a differentiable next-step image drift, a semantic intervention of strength $\alpha$ yields the local sensitivity
\begin{equation}
 \left.\frac{d x_{j+2}}{d\alpha}\right|_{\alpha=0}
 =\Delta t_{j+1}\,\frac{\partial b^x}{\partial c}\,
   \Delta t_j(2-t_j)\delta v_j^c.
\end{equation}
The Jacobian is evaluated along the unperturbed trajectory. This expression identifies the feedback channel; the FID direction is an empirical property tested by the complete sampling intervention.

\paragraph{Teacher-forced error and generated trajectories.}
The velocity probe measures risk on forward-noised data, whereas the sampler visits model-induced joint image/semantic states. The terminal image depends on the directions of successive drift changes and on subsequent sensitivity to both states. Consequently, the observed reduction in FID can accompany an increase in teacher-forced NMSE. The projector is a nonlinear MLP; its local Jacobian and the downstream network can be sensitive to different hidden-state directions.

\begin{figure}[t]
\centering\includegraphics[width=\linewidth]{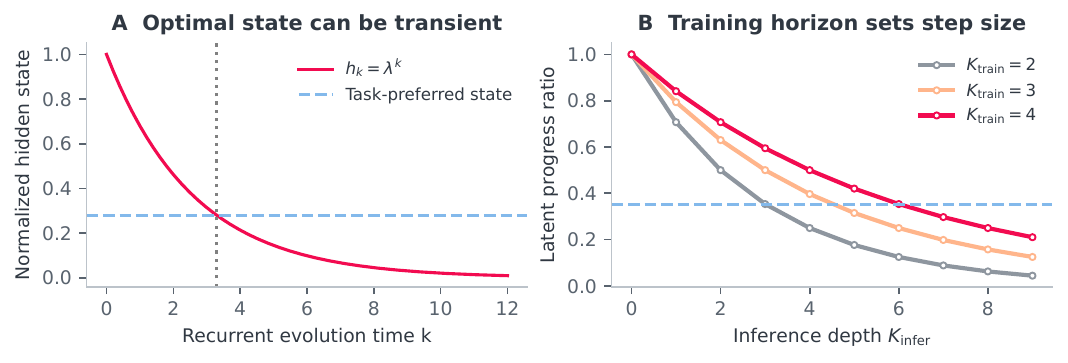}
\caption{\textbf{Transient optima and horizon-dependent latent progress.} A contractive trajectory can pass through a task-preferred state before reaching its fixed point. Under a fixed total-progress assumption, a larger training horizon corresponds to smaller recurrent steps and a later optimal inference horizon.}
\label{fig:theory}
\end{figure}

\section{Paired Output-Depth Interventions}
\label{app:causal}
\paragraph{Experimental question and replication.}
We use shared Block4 SiT-B/2 REG checkpoints trained for 400K steps at $\ktr=m\in\{2,3\}$. Mixed policies evaluate depths $m$ and $m+1$ at the same current $(x,c,t,y)$ of that condition's trajectory; equal-depth policies evaluate the network once. We select image velocity from depth $K_x$ and semantic velocity from depth $K_c$. Conditions A/B/C/D use $(m,m),(m+1,m),(m,m+1),(m+1,m+1)$ throughout sampling. Each selected output updates its corresponding state, so later inputs depend on both policies.

For train-two, the four conditions and primary directional comparison $\operatorname{FID}(B)>\operatorname{FID}(D)$ were specified before a seed-1 screening run with 5K images per condition. Seeds 2 and 3 then evaluated all four conditions at 50K each with the checkpoint, depths, and comparison fixed. Train-three was designated as a cross-horizon replication with depths three and four before its evaluation. Its two 50K seeds use the same factorial design. All 16 final evaluations are reported; screening samples are excluded from the table and means. A/D reproduce each checkpoint's native matched and one-step-extrapolated policies. This intervention tests a fixed extra step, rather than searching for the best depth of either checkpoint.

\paragraph{Sampling and common random numbers.}
All conditions use EMA, CFG=1, the original 250-update Euler--Maruyama time grid, FP32 network evaluation with TF32, FP64 state arithmetic, and the same VAE, quantization, ImageNet reference, and Inception evaluator. Train-two uses six H100 GPUs and train-three uses four, each with batch 64. Rank seeds are $sW+r$, where $s$ is the generation seed, $W$ the world size, and $r$ the rank. Within each checkpoint and seed, initial image and semantic noise, class labels, and every SDE noise increment are shared across all four conditions. Each condition is a fresh complete sampling run. Different world sizes make cross-horizon comparisons unpaired, with partially overlapping rank seeds. The two checkpoints represent different training horizons, not repeated training seeds.

\begin{table}[htbp]
\centering\small\setlength{\tabcolsep}{3pt}\begin{tabular}{lccrrrrrr}
\toprule
Condition & $K_x$ & $K_c$ & FID $\downarrow$ & sFID $\downarrow$ & IS $\uparrow$ & Prec. $\uparrow$ & Rec. $\uparrow$ & Calls \\
\midrule
\multicolumn{9}{l}{\textit{$\ktr=2$, generation seed 2; 50,000 images per condition}} \\
A & 2 & 2 & 14.709 & 6.702 & 97.933 & 0.6473 & \textbf{0.6326} & 250 \\
B & 3 & 2 & 7.533 & 6.236 & 142.720 & 0.7455 & 0.5599 & 500 \\
C & 2 & 3 & 9.362 & 6.097 & 132.561 & 0.7021 & 0.5974 & 500 \\
D & 3 & 3 & \textbf{5.339} & \textbf{5.783} & \textbf{185.375} & \textbf{0.7934} & 0.5143 & 250 \\
\midrule
\multicolumn{9}{l}{\textit{$\ktr=2$, generation seed 3; 50,000 images per condition}} \\
A & 2 & 2 & 14.567 & 6.632 & 98.236 & 0.6482 & \textbf{0.6381} & 250 \\
B & 3 & 2 & 7.398 & 6.207 & 144.263 & 0.7472 & 0.5660 & 500 \\
C & 2 & 3 & 9.175 & 6.067 & 133.129 & 0.7047 & 0.5971 & 500 \\
D & 3 & 3 & \textbf{5.296} & \textbf{5.758} & \textbf{187.057} & \textbf{0.7970} & 0.5130 & 250 \\
\midrule
\multicolumn{9}{l}{\textit{$\ktr=3$, generation seed 2; 50,000 images per condition}} \\
A & 3 & 3 & 13.924 & 6.552 & 100.534 & 0.6555 & \textbf{0.6365} & 250 \\
B & 4 & 3 & 7.438 & 5.480 & 139.488 & 0.7519 & 0.5747 & 500 \\
C & 3 & 4 & 9.279 & 6.155 & 132.510 & 0.7111 & 0.6005 & 500 \\
D & 4 & 4 & \textbf{5.235} & \textbf{5.295} & \textbf{176.790} & \textbf{0.8011} & 0.5296 & 250 \\
\midrule
\multicolumn{9}{l}{\textit{$\ktr=3$, generation seed 3; 50,000 images per condition}} \\
A & 3 & 3 & 13.848 & 6.612 & 100.628 & 0.6583 & \textbf{0.6339} & 250 \\
B & 4 & 3 & 7.519 & 5.540 & 139.223 & 0.7535 & 0.5700 & 500 \\
C & 3 & 4 & 9.303 & 6.193 & 130.924 & 0.7167 & 0.5902 & 500 \\
D & 4 & 4 & \textbf{5.339} & \textbf{5.351} & \textbf{175.520} & \textbf{0.8038} & 0.5294 & 250 \\
\bottomrule
\end{tabular}

\caption{All paired 50K output-depth evaluations. Bold marks the best metric within each checkpoint and generation seed. Calls counts denoiser evaluations per image: A/D use 250 and B/C use 500, all with 250 state updates.}
\label{tab:causal_full}
\end{table}

\paragraph{Outputs, trajectories, and compute.}
For a mixed policy, the two network calls use identical current inputs before either state is advanced. Thus B$\to$D keeps the image-update function at depth $m+1$, but changes future semantic inputs to that function. A$\to$C similarly isolates the semantic-output policy at image depth $m$. Changing image depth under a fixed semantic policy gives A$\to$B and C$\to$D. The mixed policies require 500 network calls because both depths are evaluated separately; the native joint endpoints require 250. These interventions identify output-path effects rather than provide a compute-matched alternative sampler.

\paragraph{Implementation checks.}
Routed A/D match the native sampler's image latents at their respective depths with identical inputs and random states. Mixed outputs match direct model outputs evaluated at those same inputs. Replacing only the final semantic output leaves the final image unchanged, as Proposition~\ref{prop:feedback} requires. Model forwards consume no random numbers, and the recorded native SDE draws preserve pairing.

\paragraph{FID, interaction, and coverage.}
Both checkpoints and both seeds give D$<$B$<$C$<$A in FID. For train-two, B$\to$D improves FID by \result{CausalTrain2Seed2BDGain}\% and \result{CausalTrain2Seed3BDGain}\%; train-three gives \result{CausalTrain3Seed2BDGain}\% and \result{CausalTrain3Seed3BDGain}\%. Joint refinement improves sFID, IS, and precision while lowering recall in all four groups. The FID-scale interaction $J_{11}-J_{10}-J_{01}+J_{00}$ is positive in each group: the joint FID reduction is smaller than the sum of isolated reductions. This nonlinear metric contrast does not assign additive percentages of the gain to separate mechanisms. Means average two separate 50K metrics within each checkpoint; they are not pooled 100K FIDs or estimates of training-seed uncertainty.

\section{Site Interventions, Depth Policies, and Controls}
\label{app:controls}
\label{sec:elasticity}
\subsection{Site and Supervision Depth}
Table~\ref{tab:site} supplements Figure~\ref{fig:site} with both inference depths and recall for every tested site/depth pair.
\begin{table}[t]
\centering\small\setlength{\tabcolsep}{4pt}\begin{tabular}{rrrrrrr}
\toprule
Sup. depth & Loop block & FID $K_{\mathrm{infer}}=2$ & FID $K_{\mathrm{infer}}=3$ & FID $K_{\mathrm{infer}}=4$ & Gain (\%) & R at $K_{\mathrm{infer}}=3$ \\
\midrule
\multicolumn{7}{l}{\textit{Fixed supervision at depth four}} \\
4 & 2 & \textbf{\result{D4B2K2}} & \result{D4B2K3} & \result{D4B2K4} & \result{D4B2Gain} & \result{D4B2Recall} \\
4 & 4 & \result{D4B4K2} & \textbf{\result{D4B4K3}} & \result{D4B4K4} & \result{D4B4Gain} & \result{D4B4Recall} \\
4 & 6 & \textbf{\result{D4B6K2}} & \result{D4B6K3} & \result{D4B6K4} & \result{D4B6Gain} & \result{D4B6Recall} \\
4 & 8 & \textbf{\result{D4B8K2}} & \result{D4B8K3} & \result{D4B8K4} & \result{D4B8Gain} & \result{D4B8Recall} \\
\midrule
\multicolumn{7}{l}{\textit{Supervision and recurrence moved together}} \\
2 & 2 & \result{D2B2K2} & \textbf{\result{D2B2K3}} & \result{D2B2K4} & \result{D2B2Gain} & \result{D2B2Recall} \\
4 & 4 & \result{D4B4K2} & \textbf{\result{D4B4K3}} & \result{D4B4K4} & \result{D4B4Gain} & \result{D4B4Recall} \\
6 & 6 & \result{D6B6K2} & \textbf{\result{D6B6K3}} & \result{D6B6K4} & \result{D6B6Gain} & \result{D6B6Recall} \\
8 & 8 & \result{D8B8K2} & \textbf{\result{D8B8K3}} & \result{D8B8K4} & \result{D8B8Gain} & \result{D8B8Recall} \\
\bottomrule
\end{tabular}

\caption{\textbf{Site/depth interventions.} Bold marks the lowest FID within each row. Gain is the relative FID reduction from $\kinfer=2$ to $\kinfer=3$; negative values indicate degradation. The last column reports recall at the extrapolated depth. The d4/B4 anchor is repeated across groups. At d8/B8, FID improves while sFID worsens.}
\label{tab:site}
\end{table}

\subsection{Training and Inference Depth Policies}
Training over different horizons controls the range of useful inference depths. LoopREG-x3 at $\kinfer=5$ has precision \result{XthreeBestprecision} and recall \result{XthreeBestrecall}, compared with the REG baseline's \result{REGbaseprecision} and \result{REGbaserecall}. Lower FID therefore coexists with reduced measured coverage. RandomLoop broadens the region of useful depths and retains recall closer to the baseline, at a worse peak FID. For example, Random $\{1,2,3\}$ attains its best observed FID of \result{Random123fid} at $\kinfer=\result{Random123infer_k}$ with recall \result{Random123recall}. Fixed depth favors peak FID; random depth favors a broader useful range. Some random-policy curves continue improving at the evaluated boundary.

A timestep-dependent depth schedule gives another operating point. Using $\kinfer=2$ for $t<0.6$ and $\kinfer=3$ for $t\geq0.6$ yields FID \result{Adaptivefid} and recall \result{Adaptiverecall}, between fixed $\kinfer=2$ and fixed $\kinfer=3$. Its mean recurrent depth is \result{AdaptiveMeanK}; relative to fixed $\kinfer=3$, recurrent-core executions decrease by \result{AdaptiveCoreSavings}\%, and total block executions by only \result{AdaptiveBlockSavings}\%. Neither quantity is a measured wall-clock or full-model FLOP saving.

\begin{figure}[htbp]
\centering\includegraphics[width=\linewidth]{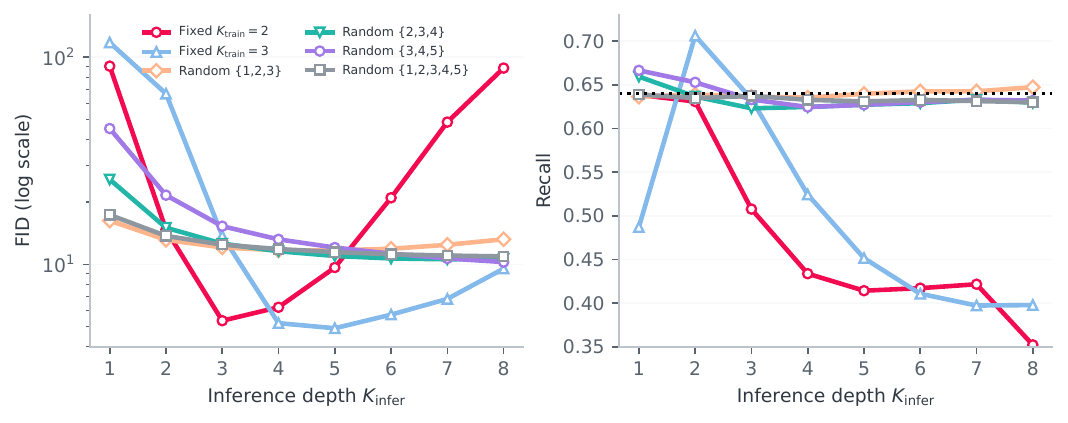}
\caption{Fixed and random training horizons. The full observed depth range exposes under-roll failures, finite optima, and recall differences. The dotted recall line is the local REG baseline. Random $\{3,4,5\}$ and $\{1,2,3,4,5\}$ are still improving in FID at $\kinfer=8$, the evaluated boundary.}
\end{figure}

\subsection{Shared and Periodic Unshared Operators}
For an unshared pair, $H=F_b\circ F_a$ and $\kinfer=2m$ evaluates $H^m$; for a triplet, $H=F_c\circ F_b\circ F_a$ and $\kinfer=3m$. Inference cycles through the trained copies without adding weights. The matched pair attains FID \result{UnsharedK2}, slightly better than shared train-two's \result{REGtwofid}. It improves to \result{UnsharedK4} on its second cycle before deteriorating sharply. The triplet worsens from \result{UnsharedThreeK3fid} to \result{UnsharedThreeK6fid} on its first extra cycle. Thus sharing does not explain a matched-depth advantage; its main benefit here is greater useful extrapolation.
\begin{table}[htbp]
\centering\small\setlength{\tabcolsep}{3pt}\begin{tabular}{llrrrrrr}
\toprule
$K_{\mathrm{train}}$ / Policy & $K_{\mathrm{infer}}$ & Blocks & FID $\downarrow$ & sFID $\downarrow$ & IS $\uparrow$ & P $\uparrow$ & R $\uparrow$ \\
\midrule
2 / fixed & 1 & 12 & 151.9461 & 50.3472 & 6.8845 & 0.1480 & 0.1539 \\
2 / fixed & 2 & 13 & \textbf{14.1520} & \textbf{6.6645} & \textbf{99.3560} & \textbf{0.6541} & \textbf{0.6383} \\
\midrule
2 / periodic & 2 & 13 & 14.1520 & \textbf{6.6645} & 99.3560 & \textbf{0.6541} & \textbf{0.6383} \\
2 / periodic & 4 & 15 & \textbf{12.2748} & 8.2114 & \textbf{137.2766} & 0.6315 & 0.5338 \\
2 / periodic & 6 & 17 & 61.1025 & 19.6300 & 39.9223 & 0.2887 & 0.4964 \\
2 / periodic & 8 & 19 & 147.5555 & 52.2028 & 10.3376 & 0.0813 & 0.2113 \\
\midrule
3 / periodic & 3 & 14 & \textbf{13.3211} & \textbf{6.5114} & \textbf{103.6684} & \textbf{0.6622} & \textbf{0.6294} \\
3 / periodic & 6 & 17 & 40.9129 & 23.2540 & 53.8996 & 0.4237 & 0.5638 \\
\bottomrule
\end{tabular}

\caption{\textbf{Unshared controls at 400K.} Every evaluated depth is included, grouped by training horizon and inference policy. The fixed and periodic pair share their matched-depth anchor. Bold marks metric optima within each group.}
\label{tab:unshared_final}
\end{table}

\subsection{Teacher-Forced Velocity Error}
\begin{table}[htbp]
\centering\small\setlength{\tabcolsep}{3pt}\begin{tabular}{lrrrrr}
\toprule
Model & $K_{\mathrm{train}}$ & NMSE $K_{\mathrm{infer}}$ & FID $K_{\mathrm{infer}}$ & NMSE train & NMSE FID-best \\
\midrule
LoopSiT-x2 & 2 & 2 & 2 & \result{LoopSiTx2NMSE2} & -- \\
LoopREG-x2 & 2 & 2 & 3 & \result{LoopREGx2NMSE2} & \result{LoopREGx2NMSE3} \\
LoopREG-x3 & 3 & 3 & 5 & \result{LoopREGx3NMSE3} & \result{LoopREGx3NMSE5} \\
\bottomrule
\end{tabular}

\caption{The separate N=256 teacher-forced velocity probe, averaged over five noise times. The FID-best horizon differs from the NMSE-best horizon for both LoopREG checkpoints. ``--'' means that the two horizons coincide.}
\end{table}
For LoopREG, velocity NMSE is lowest at the training horizon, whereas generation FID is lowest at a larger depth. The extra-depth benefit therefore cannot be reduced to improvement in this measured one-step teacher-forced error.

\section{Depth-Selection Diagnostics}
\label{app:depthselection}
The main method selects inference depth from measured sweeps. We additionally tested two candidate forecasting rules on the existing train-two, train-three, and train-four checkpoints. Their historical endpoint behavior was known during development; predictions were frozen before new evaluation samples were generated. These diagnostics use 5K or 4,096 images per metric, separately from the main 50K results.

\paragraph{Early training curves.}
Exponential fits $E_q(s)=E_{q,\infty}+A_q e^{-\gamma_q s}$ model FID's mean and covariance contributions from observations at 50K, 100K, and 150K steps. The fitted rate must be positive and the limit nonnegative; invalid fits abstain. The grid contains three training horizons and six inference depths, $\kinfer\in\{2,\ldots,7\}$. Calibration uses seed 1701; 200K and 400K evaluations use seed 2901 with paired streams and 5K samples per cell.
\begin{table}[htbp]
\centering\small\setlength{\tabcolsep}{4pt}\begin{tabular}{lrrrr}
\toprule
Forecast & MAE, 200K & MAE, 400K & Depth hits, 400K & Selected $(\ktr,\kinfer)$ \\
\midrule
Component exponential & 0.638 & 3.262 & 1/3 & (3,5) \\
Total-FID exponential & 0.652 & 3.299 & 1/3 & (3,5) \\
Latest observation (150K) & 2.237 & 5.258 & 1/3 & (3,5) \\
\bottomrule
\end{tabular}

\caption{Early-curve forecasts on 5K-sample FID. MAE averages 18 candidates. Depth hits count correct inference-depth selections across three trained models.}
\label{tab:forecast_r4}
\end{table}
All three rules select the lowest-FID pair $(\ktr,\kinfer)=(3,5)$, but their per-model selections $4,5,7$ match only one of the observed optima $3,5,6$. Exponential fitting improves numerical forecasts without improving depth selection over the 150K ranking. At 400K, \result{R4BelowFloor} of 18 measured FIDs fall below their fitted limits. This rule requires early trained checkpoints and does not select training depth before training.

\paragraph{Local sampling responses.}
The second rule interpolates both image and semantic velocities as $\hat v_\alpha=(1-\alpha)\hat v_k+\alpha\hat v_{k+1}$. Complete sampling at $\alpha\in\{0,1/16,1/8,1/4\}$ produces terminal Inception features after decoding and quantization. Linear and quadratic fits extrapolate these features to $\alpha=1$, using 4,096 paired samples at calibration seed 5101. Predictions are fixed before independent-seed endpoint evaluation at seed 5201. The eight edges are $2\to3,3\to4$ for train-two, $3\to4,4\to5,5\to6$ for train-three, and $4\to5,5\to6,6\to7$ for train-four.
\begin{table}[htbp]
\centering\small\setlength{\tabcolsep}{4pt}\begin{tabular}{lrrrr}
\toprule
Method & Direction hits & Depth hits & Change MAE & Mean selection loss \\
\midrule
Quadratic feature extrapolation & 3/8 & 0/3 & 8550.500 & 8.744 \\
Quadratic, half step & 3/8 & 0/3 & 82100.946 & 8.744 \\
Linear feature extrapolation & 3/8 & 0/3 & 242.900 & 8.744 \\
Zero change & 0/8 & 0/3 & 3.560 & 8.744 \\
$\ktr+1$ & -- & 1/3 & -- & 0.405 \\
Observed calibration anchors & -- & 3/3 & -- & 0.000 \\
\bottomrule
\end{tabular}

\caption{Local feature extrapolation on 4,096-sample FID. Direction hits compare signs over eight edges. Selection loss is measured FID at the selected depth minus the best measured FID; depths are chosen by accumulated predicted changes, with smaller-depth tie-breaking.}
\label{tab:forecast_r5}
\end{table}
Endpoint evaluations again favor depths $3,5,6$. All three nonzero extrapolators predict deterioration on every edge, missing the five beneficial edges and selecting the trained depths. Their mean selection loss is \result{R5MeanRegret} FID. The $\kinfer=\ktr+1$ rule performs better; ranking observed calibration anchors recovers all three optima using its multiple-depth measurements. The tested local feature approximations therefore do not supply a predictive stopping rule.

\paragraph{Earlier representation diagnostics.}
\label{app:theoryprobes}
A development score combining gradient compatibility, geometry gain, and perturbation response selects co-located depth two, whereas depth four yields the best measured FID. All four reported semantic-direction perturbation ratios exceed one. A median geometry-peak rule predicts horizons $3,4,6$, compared with observed $3,5,6$. A separate frozen REG-baseline score selects its already supervised block four while its mean extra-step geometry gains remain negative at every site. These results distinguish supervision placement and recurrent training from prediction of the optimal depth.

\section{XL Training Milestones}
\label{app:xl}
Table~\ref{tab:xl_milestones} summarizes the 50K, 100K, 200K, and 400K checkpoints, together with the 220K checkpoint used in the main-text comparison with published REG. Each row identifies the evaluated inference-depth set; a dash marks a checkpoint evaluated at only one depth. Gain is relative to the same checkpoint at $\kinfer=2$, and recall is reported at its best observed FID.
\begin{table}[htbp]
\centering\small\setlength{\tabcolsep}{3pt}\begin{tabular}{rlrrrrr}
\toprule
Steps & Tested $K_{\mathrm{infer}}$ & FID at 2 & Best $K_{\mathrm{infer}}$ & Best FID & Gain (\%) & Recall \\
\midrule
50K & 2,3,4,5 & 25.1962 & 4 & \textbf{11.8154} & 53.11 & 0.3407 \\
100K & 2,3,4,5 & 8.3360 & 3 & \textbf{4.7552} & 42.96 & 0.4682 \\
200K & 2,3,4,5 & 4.6383 & 3 & \textbf{3.4942} & 24.67 & 0.5010 \\
220K & 3 & -- & -- & 3.3675 & -- & 0.5180 \\
400K & 1,2,3,4,5 & 3.3823 & 3 & \textbf{2.8376} & 16.10 & 0.5490 \\
\bottomrule
\end{tabular}

\caption{\textbf{Representative XL training checkpoints.} Best depths are minima over the listed sets. The 220K checkpoint supplies a single-depth observation. All metrics at the final 400K checkpoint appear in Appendix~\ref{app:sweeps}.}
\label{tab:xl_milestones}
\end{table}
Published REG results are reference values from Table~9 of \citet{reg}; the comparison uses no local non-loop REG-XL run. Checkpoint crossings reported in the main text refer to the earliest evaluated observations in the full record, rather than exact crossing times.

\clearpage
\section{DINO Representation Trajectories}
\label{app:dynamics}
The final probe evaluates 1,024 forward-noised real images at $t\in\{.1,.3,.5,.7,.9\}$. Unbiased linear CKA \citep{cka} compares mean hidden patch features with either mean DINO patch features or DINO CLS features. The launcher caps each CKA calculation at a deterministic 256-image subset; this cap is specified by the launcher rather than the saved per-model metadata. ``CLS CKA'' names the teacher target, not the model's semantic-token state. Spatial correlations compare off-diagonal normalized patch Gram entries, while native cosine compares normalized projected features with DINO targets. Both use all 1,024 images and 64 sampled patch positions.

\begin{figure}[htbp]
\centering\includegraphics[width=\linewidth]{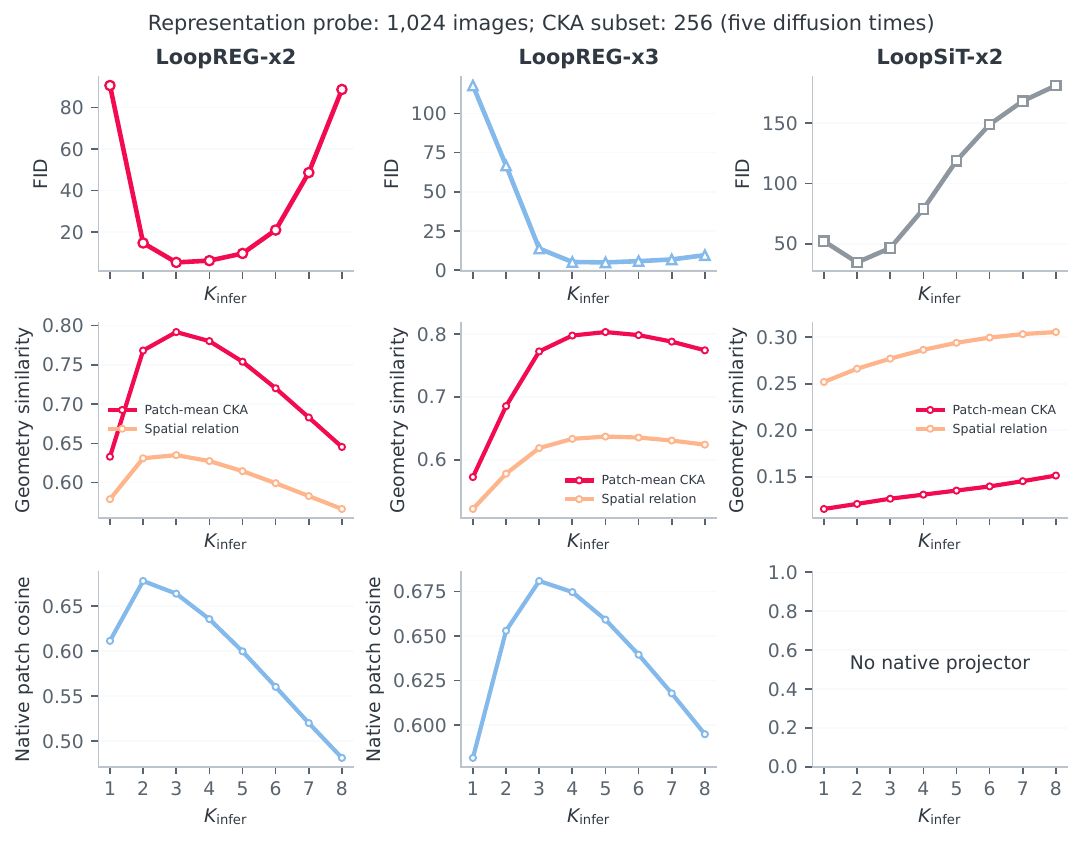}
\caption{\textbf{Representation geometry and native alignment along recurrent depth.} FID, patch-mean CKA/spatial correlation, and native projector cosine use aligned depth axes. Geometry is averaged over five noise times from the final 1,024-image probe (CKA subset: 256). These are teacher-forced hidden trajectories; FID comes from separate full-generation evaluations. Vertical scales differ across panels.}
\label{fig:geometry}
\end{figure}

For LoopREG trained at $\ktr=2$, patch-mean CKA and spatial correlation peak at $\kinfer=3$, matching the FID minimum; native projector cosine peaks at $\kinfer=2$. For $\ktr=3$, all three geometry metrics peak at $\kinfer=5$, while native cosine peaks at $\kinfer=3$. The correspondence depends on the metric and model: teacher-CLS CKA for train-two peaks at four, and LoopSiT's geometry similarity increases even as FID deteriorates. Relative hidden-state increments also do not establish global contraction. The full probe covers five models, eight recurrent depths, and five diffusion times, including baseline hidden-state probes. Those baseline probes are not generation results at additional depths.

\clearpage
\section{Final-Checkpoint Generation Sweeps}
\label{app:sweeps}
The following tables give every evaluated inference depth at the 400K checkpoint for the core SiT/REPA/REG comparisons and both larger scales. This selection is by model and checkpoint, not by metric value. Site interventions, random-depth policies, and unshared controls are summarized in Appendix~\ref{app:controls}. 

``Blocks'' counts Transformer-block executions, not exact FLOPs. Bold marks each metric's optimum within a model and checkpoint; FID and sFID are minimized, and IS, precision, and recall are maximized. Displayed values are rounded; canonical records retain source precision.
\subsection{SiT-B/2: Supervision and Training Horizon}
\begingroup\small\setlength{\tabcolsep}{5pt}
\begin{longtable}{rrrrrrr}
\toprule
$K_{\mathrm{infer}}$ & Blocks & FID $\downarrow$ & sFID $\downarrow$ & IS $\uparrow$ & P $\uparrow$ & R $\uparrow$ \\
\midrule
\endfirsthead
\toprule
$K_{\mathrm{infer}}$ & Blocks & FID $\downarrow$ & sFID $\downarrow$ & IS $\uparrow$ & P $\uparrow$ & R $\uparrow$ \\
\midrule
\endhead
\midrule
\multicolumn{7}{r}{\textit{Continued on the next page}} \\
\endfoot
\bottomrule
\endlastfoot
\rowcolor{black!5}\multicolumn{7}{l}{\textbf{LoopSiT-B/2, B4, $K_{\mathrm{train}}=2$}} \\*
1 & 12 & 52.0813 & 11.3411 & 28.9425 & 0.3676 & \textbf{0.6608} \\*
2 & 13 & \textbf{34.4524} & \textbf{6.4313} & \textbf{43.0118} & \textbf{0.5334} & 0.6387 \\*
3 & 14 & 46.4096 & 10.0098 & 34.2806 & 0.4637 & 0.6182 \\*
4 & 15 & 78.8033 & 19.1662 & 20.7103 & 0.3153 & 0.5667 \\*
5 & 16 & 118.9361 & 31.6541 & 12.4613 & 0.1959 & 0.4317 \\*
6 & 17 & 149.0437 & 42.9569 & 8.6622 & 0.1284 & 0.2810 \\*
7 & 18 & 168.4506 & 53.1676 & 6.6430 & 0.0944 & 0.1592 \\*
8 & 19 & 181.3879 & 63.7429 & 5.4062 & 0.0754 & 0.0928 \\
\midrule
\rowcolor{black!5}\multicolumn{7}{l}{\textbf{LoopREPA-B/2, d=4, B4, $K_{\mathrm{train}}=2$}} \\*
1 & 12 & 88.8098 & 20.6534 & 14.7986 & 0.2660 & 0.6509 \\*
2 & 13 & 21.9274 & \textbf{6.3655} & 66.5780 & 0.6013 & \textbf{0.6515} \\*
3 & 14 & \textbf{16.7427} & 6.6802 & \textbf{84.6709} & \textbf{0.6285} & 0.6147 \\*
4 & 15 & 28.3614 & 9.0526 & 61.8826 & 0.4946 & 0.6118 \\*
5 & 16 & 59.4920 & 14.8863 & 30.6378 & 0.2966 & 0.5953 \\*
6 & 17 & 101.3782 & 27.0815 & 15.0184 & 0.1545 & 0.4585 \\*
7 & 18 & 134.6486 & 45.3181 & 9.2247 & 0.0925 & 0.2664 \\*
8 & 19 & 158.8770 & 64.4720 & 6.6918 & 0.0646 & 0.1125 \\
\midrule
\rowcolor{black!5}\multicolumn{7}{l}{\textbf{LoopREG-B/2, d=4, B4, $K_{\mathrm{train}}=2$}} \\*
1 & 12 & 90.5972 & 16.9459 & 14.6801 & 0.2443 & \textbf{0.6385} \\*
2 & 13 & 14.7094 & 6.7629 & 97.1320 & 0.6496 & 0.6309 \\*
3 & 14 & \textbf{5.3434} & \textbf{5.8021} & 185.9796 & 0.7958 & 0.5077 \\*
4 & 15 & 6.2050 & 6.6180 & \textbf{207.7940} & \textbf{0.8078} & 0.4337 \\*
5 & 16 & 9.6552 & 9.4030 & 174.6141 & 0.7287 & 0.4142 \\*
6 & 17 & 20.9611 & 15.3694 & 108.8510 & 0.5695 & 0.4171 \\*
7 & 18 & 48.6352 & 25.9636 & 50.0054 & 0.3735 & 0.4216 \\*
8 & 19 & 88.6875 & 40.5806 & 22.4223 & 0.2294 & 0.3521 \\
\midrule
\rowcolor{black!5}\multicolumn{7}{l}{\textbf{LoopREG-B/2, d=4, B4, $K_{\mathrm{train}}=3$}} \\*
1 & 12 & 117.7844 & 33.5290 & 10.0884 & 0.1792 & 0.4871 \\*
2 & 13 & 66.7103 & 16.5472 & 23.7143 & 0.3315 & \textbf{0.7063} \\*
3 & 14 & 13.7500 & 6.6929 & 100.8156 & 0.6578 & 0.6337 \\*
4 & 15 & 5.2002 & \textbf{5.4001} & 178.3913 & 0.8023 & 0.5243 \\*
5 & 16 & \textbf{4.9136} & 5.4032 & 217.6060 & 0.8506 & 0.4515 \\*
6 & 17 & 5.7158 & 5.9812 & \textbf{224.6076} & \textbf{0.8513} & 0.4107 \\*
7 & 18 & 6.8081 & 7.1051 & 209.2107 & 0.8195 & 0.3972 \\*
8 & 19 & 9.5347 & 8.9836 & 173.4586 & 0.7508 & 0.3977 \\
\midrule
\rowcolor{black!5}\multicolumn{7}{l}{\textbf{LoopREG-B/2, d=4, B4, $K_{\mathrm{train}}=4$}} \\*
1 & 12 & 121.6970 & 35.4991 & 9.3515 & 0.1827 & 0.4151 \\*
2 & 13 & 93.8612 & 23.3516 & 14.4801 & 0.2425 & 0.6437 \\*
3 & 14 & 44.1952 & 11.7995 & 39.1530 & 0.4310 & \textbf{0.7128} \\*
4 & 15 & 13.2435 & 6.5524 & 102.6734 & 0.6605 & 0.6335 \\*
5 & 16 & 5.8749 & 5.3505 & 161.3391 & 0.7849 & 0.5518 \\*
6 & 17 & \textbf{5.1874} & \textbf{5.2995} & 193.9240 & 0.8394 & 0.4845 \\*
7 & 18 & 5.8128 & 5.6755 & \textbf{206.5647} & \textbf{0.8531} & 0.4439 \\*
8 & 19 & 6.6785 & 6.2577 & 204.0657 & 0.8496 & 0.4178 \\
\end{longtable}
\endgroup
\subsection{SiT-L/2 and SiT-XL/2}
\begingroup\small\setlength{\tabcolsep}{5pt}
\begin{longtable}{rrrrrrr}
\toprule
$K_{\mathrm{infer}}$ & Blocks & FID $\downarrow$ & sFID $\downarrow$ & IS $\uparrow$ & P $\uparrow$ & R $\uparrow$ \\
\midrule
\endfirsthead
\toprule
$K_{\mathrm{infer}}$ & Blocks & FID $\downarrow$ & sFID $\downarrow$ & IS $\uparrow$ & P $\uparrow$ & R $\uparrow$ \\
\midrule
\endhead
\midrule
\multicolumn{7}{r}{\textit{Continued on the next page}} \\
\endfoot
\bottomrule
\endlastfoot
\rowcolor{black!5}\multicolumn{7}{l}{\textbf{LoopREG-L/2, d=8, B8, $K_{\mathrm{train}}=2$}} \\*
1 & 24 & 59.4139 & 12.6041 & 26.0546 & 0.3214 & \textbf{0.7664} \\*
2 & 25 & 4.2286 & 5.0924 & 170.3622 & 0.7516 & 0.6420 \\*
3 & 26 & \textbf{3.2082} & \textbf{4.4353} & 262.6577 & 0.8593 & 0.5201 \\*
4 & 27 & 5.4971 & 4.5898 & \textbf{291.8665} & \textbf{0.8827} & 0.4438 \\*
5 & 28 & 7.0317 & 4.9216 & 267.8294 & 0.8450 & 0.4206 \\
\midrule
\rowcolor{black!5}\multicolumn{7}{l}{\textbf{LoopREG-XL/2, d=8, B8, $K_{\mathrm{train}}=2$}} \\*
1 & 28 & 52.4630 & 10.6201 & 31.5308 & 0.3699 & \textbf{0.7668} \\*
2 & 29 & 3.3823 & 4.8470 & 184.7094 & 0.7584 & 0.6347 \\*
3 & 30 & \textbf{2.8376} & \textbf{4.6353} & \textbf{254.6991} & \textbf{0.8386} & 0.5490 \\*
4 & 31 & 3.7734 & 5.1102 & 239.4408 & 0.8179 & 0.5240 \\*
5 & 32 & 7.8614 & 7.0541 & 164.5297 & 0.7154 & 0.5457 \\
\end{longtable}
\endgroup

\end{document}